\pdfoutput=1
\documentclass[11pt]{article}

\usepackage[final]{acl}
\usepackage{times}
\usepackage{latexsym}
\usepackage{amsmath}
\usepackage{cleveref}
\usepackage{multirow}
\usepackage{booktabs}
\usepackage{tabularx}
\usepackage{array}
\usepackage{makecell}
\usepackage{xfrac}
\usepackage{xcolor}
\usepackage{colortbl}
\usepackage{hhline}

\usepackage{caption}
\usepackage{subcaption}
\usepackage[T1]{fontenc}
\usepackage[utf8]{inputenc}

\usepackage{microtype}

\usepackage{inconsolata}

\usepackage{graphicx}
\usepackage{hyperref}

\title{Towards a Client-Centered Assessment of LLM Therapists by\\Client Simulation}
\author{Jiashuo Wang$^{1}$\quad Yang Xiao$^{1}$\quad Yanran Li$^{2}$\quad Changhe Song$^{1}$\quad Chunpu Xu$^{1}$\\\textbf{Chenhao Tan}$^{3}$\quad \textbf{Wenjie Li}$^{1}$ \\
$^{1}$The Hong Kong Polytechnic University \quad $^{2}$Independent Researcher \\ \quad $^{3}$University of Chicago \\
\small{\texttt{\{csjwang,cswjli\}@comp.polyu.edu.hk} \quad \texttt{\{yang-alan.xiao,changhe.song,chun-pu.xu\}}@connect.polyu.hk}\\
\small{\texttt{yanranli.summer@gmail.com} \quad \texttt{chenhao@chenhaot.com}}}

\begin{document}
\maketitle
\begin{abstract}
\textcolor{red}{This work does \textbf{NOT} advocate for the use of large language models (LLMs) in therapy. Instead, it proposes an \textit{assessment} approach to \textit{reveal the characteristics} of LLM therapists.}

Although there is a growing belief that LLMs can be used as therapists, exploring LLMs' capabilities and inefficacy, particularly from the client's perspective, is limited.
This work focuses on a client-centered assessment of LLM therapists with the involvement of simulated clients, a standard approach in clinical medical education.
However, there are two challenges when applying the approach to assess LLM therapists at scale.
Ethically, asking humans to frequently mimic clients and exposing them to potentially harmful LLM outputs can be risky and unsafe. 
Technically, it can be difficult to consistently compare the performances of different LLM therapists interacting with the same client.
To this end, we adopt LLMs to simulate clients and propose ClientCAST, a client-centered approach to assessing LLM therapists by client simulation. 
Specifically, the simulated client is utilized to interact with LLM therapists and complete questionnaires related to the interaction. 
Based on the questionnaire results, we assess LLM therapists from three client-centered aspects: session outcome, therapeutic alliance, and self-reported feelings. 
We conduct experiments to examine the reliability of ClientCAST and use it to evaluate LLMs therapists implemented by Claude-3, GPT-3.5, LLaMA3-70B, and Mixtral 8\text{\texttimes}7B.\footnote{Codes are released at \url{https://github.com/wangjs9/ClientCAST}.}
\end{abstract}

\begin{figure*}[ht]
\centering
\includegraphics[width=\linewidth]{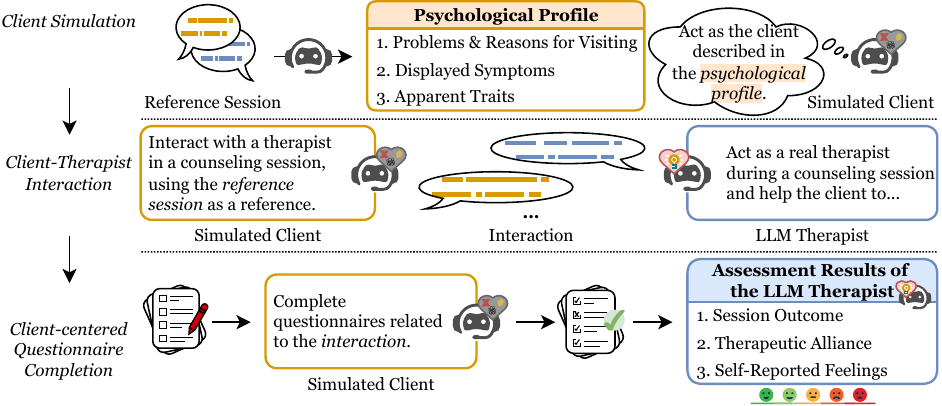} 
\caption{This is the overview framework of ClientCAST.
It utilizes an LLM as a simulated client equipped with a specific psychological profile. 
The simulated client interacts with an LLM therapist and completes questionnaires regarding their interaction. 
Finally, ClientCAST provides a client-centered assessment of the LLM therapist based on the results of questionnaires.}
\label{fig:clientcast}
\end{figure*}
\section{Introduction}
Ever since ELIZA, a therapy chatbot, was found to provide emotional support, it has been argued that chatbots could scale up mental health support \cite{colby1966computer}.
Recently, the advanced abilities of LLMs have further enhanced the credibility of this argument, evident by both research studies \cite{nie2022conversational,hauser2022promise} and LLM end users \cite{youtube_therapist}.

While numerous users have expressed that LLM therapists helped them \cite{scientificamerican2023}, a few potential harms have been perceived \cite{de2023benefits,Xiang2023}.
To examine the capabilities and inefficacy of LLMs as therapists, several previous studies have evaluated and analyzed their behaviors \cite{chiu2024computational,li2024automatic}. 
However, these evaluations primarily focus on the therapists' perspectives, which may differ from those of the clients \cite{li2024automatic,duncan2003session,bachelor1999therapeutic}. 
Our work aims to assess LLM therapists from the clients' perspectives. 

For a client-centered assessment, it is critical to involve clients.
In clinical medical education, ``actors'' are hired and trained to simulate clients and interact with therapists for assessment purposes \cite{kuehne2018standardized}.
However, unlike in medical education, assessing LLM therapists requires scalability in client simulation.
Beyond hiring and training costs, there are ethical and technical challenges in using this approach to assess LLM therapists.
Ethically, long-term mimicking of client symptoms can cause discomfort for individuals \cite{bokken2006impact}; meanwhile, this method exposes individuals to potentially harmful LLM outputs. 
Technically, human behaviors can vary over time and across interactions, making it difficult to consistently compare the performances of different LLM therapists when interacting with the same client.

In this paper, we design ClientCAST, a \underline{Client}-\underline{C}entered approach to \underline{AS}sessing LLM \underline{T}herapists by client simulation, displayed in \Cref{fig:clientcast}.
An LLM is equipped with a specific psychological profile to simulate a client and interact with an LLM therapist, i.e., the evaluation target.
If the simulated client behaves appropriately, the LLM-simulated client allows for the client's involvement in the LLM therapist assessment while addressing the challenges mentioned above.
After the interaction, the simulated client completes questionnaires related to this interaction, i.e., the counseling session. 
These questionnaires include items that ask the client about their feelings and thoughts, focusing on three key aspects: session outcome, therapeutic alliance, and self-reported feelings.
The client-centered assessment of the LLM therapist is derived from the results of these questionnaires.

To prove the reliability of ClientCAST, we have conducted experiments using two human-human therapy counseling datasets, High-Low Quality Counseling \cite{perez2019makes} and AnnoMI \cite{wu2022anno}, and four LLMs, i.e., Claude-3, GPT-3.5, Mixtral 8\text{\texttimes}7B and LLaMA3-70B, to simulate clients.
Our findings indicate that simulated clients consistently adhere to provided problems, reasons for visiting, symptoms and speech tone, but they are less accurate in reproducing defined apparent traits.
In general, more advanced models yield more accurate simulations.
Additionally, high- and low-quality counseling sessions can be distinguished based on the completed questionnaires.
Finally, we use ClientCAST to evaluate therapists implemented by Claude-3, GPT-3.5, LLaMA3-70B, and Mixtral 8\text{\texttimes}7B.

The contributions of our work are summarized as follows: 
\textbf{(1)} We propose a novel approach to assess LLM therapists from the clients' perspectives by introducing ClientCAST, which involves LLM-simulated clients in the assessment process. 
\textbf{(2)} Through experiments, we find that simulated clients are generally consistent with their provided psychological profiles, though not perfectly, and can effectively distinguish between high- and low-quality counseling sessions.
\textbf{(3)} Utilizing ClientCAST, we assess the performance of therapists implemented by different LLMs.

\section{Related Work}
\subsection{Computational Therapist Assessment}
Previous studies evaluating LLM therapists focus primarily on therapist behaviors. 
\citet{chen2023llm} compare the types of questions and empathy behaviors of LLM therapists with those of human therapists. 
BOLT \cite{chiu2024computational} is designed to analyze the distribution and frequency of LLM therapists' behaviors, such as reflections and solutions. 
Notably, \citet{li2024automatic} use LLMs to evaluate the therapeutic alliance in human-human counseling sessions and compare the results with client and therapist evaluations. Results demonstrate that the correlation between LLM and human evaluations is higher than between clients and therapists.

\subsection{Client Simulation}
Simulated clients, aka standardized patients, are widely used in clinical medical education \cite{kuehne2018standardized,mesquita2010developing}.
An ``actor'' is instructed to behave as a client based on a provided client profile and interact with the healthcare professionals.
However, its university is limited by the high costs of hiring, training, and managing actors \cite{hillier2020standardization,barrows1993overview}.
Moreover, simulating clients can cause physical and psychological discomfort for actors consistently mimicking actual clients \cite{bokken2006impact,bokken2004performance}.
Fortunately, recent advancements in LLMs enable fair role-playing.
Prompted with appropriate client profiles, LLMs can interact as clients with expected tones and symptoms, as recent work shows:
\citet{chen2023llm} investigate the influence of various profile contents and distinct prompts on client and therapist simulation in counseling therapy.
In clinical medical scenarios, \citet{li2024leveraging} utilize prompts and RAG to guide LLMs in displaying appropriate symptoms when interacting with doctors.

\section{Preliminary}
\subsection{Counseling Therapy Datasets}
In this work, we use two datasets to conduct the experiments: \textbf{High-Low Quality Counseling} \cite{perez2019makes} and \textbf{AnnoMI} \cite{wu2022anno}.
Both datasets consist of counseling therapy transcripts carefully derived from publicly available videos on online platforms (YouTube and Vimeo). 
These transcripts maintain high standards while ensuring the confidentiality of sensitive personal information. 
Furthermore, each transcript is assessed and categorized as either a high- or low-quality counseling session based on MI psychotherapy principles. 
We use both datasets in compliance with the appropriate licenses and consents the authors provide within their terms of use.
Specifically, we utilize these two datasets to simulate clients (\Cref{sec:method:client_simulation}) and to validate the client simulation (\Cref{sec:valid:client_simulation}).
Notably, we exclude the transcripts that either lack sufficient conversational context (such as return visits without enough background information) or have limited content (with only a few conversation turns).
Finally, this work uses 213 high- and 87 low-quality sessions. 

\subsection{Involved LLMs}
We use recent popular LLMs to simulate clients in ClientCAST (\Cref{sec:method:client_simulation}) and evaluate their capabilities as therapists (\Cref{sec:evaluation}).
The models considered are Claude-3 (Haiku), GPT3.5 (GPT-3.5-turbo-0613), LLaMA3-70B and Mixtral 8\text{\texttimes}7B.

\section{ClientCAST Design}
\Cref{fig:clientcast} presents an overview of the ClientCAST framework. It consists of two essential processes: client simulation and client-centered questionnaire completion. During these processes, there is an interaction between the simulated client and the LLM therapist being assessed.

\subsection{Client Simulation}
\subsubsection{Psychological Profile}
The psychological profile provided to LLMs should significantly influence the counseling session and facilitate the identification of the specific client. 
Inspired by \citet{schneider2000development} and \citet{chen2023llm}, we consider the following information:
\textbf{(1) Problems \& Reasons for Visiting.}
We use two sentences to describe the client's problems and the reasons for visiting, respectively.
\textbf{(2) Displayed Symptoms.}
We consider 61 potential client symptoms, covering areas of depression, anxiety, symptom distress, interpersonal relations, and social roles. These symptoms are from three widely used questionnaires: PHQ-9, GAD-7, and OQ-45.
\textbf{(3) Apparent Traits.}
Apparent traits significantly affect the client's speech tone and conversation engagement. We account for the big five personality traits, emotion fluctuations (EF), unwillingness to express emotions (UWE), and resistance toward the therapist (RT). Each trait can be described at a severe level. There are five levels for the big five and three for the other traits.
More details and the psychological profile example are in \Cref{sec:appendix:client_information}.

\subsubsection{Simulation Method}\label{sec:method:client_simulation}
We formulate the client simulation as follows. Let $S_i$ represent the transcript of a counseling session involving a therapist $T_i$ and a client $C_i$. 
Our objective is to simulate $C_i$ by prompting an LLM. 
First, we extract the psychological profile $\mathcal{P}_{C}(S_i)$ of $C_i$ from $S_i$ using the LLM. 
The LLM is then instructed to behave like $C_i$ based on $\mathcal{P}_{C}(S_i)$. 
The session $S_i$ is also supplied as a reference session. 
The LLM is tasked with learning the client's speech tone and conversational style from $S_i$ but engaging in a new session with a different therapist, as if in a parallel universe. 
Prompts are in \Cref{sec:appendix:client_simulation}.

\subsection{Client-centered Questionnaire Completion}
\subsubsection{Assessment Aspects \& Questionnaires}
Inspired by \citet{mallinckrodt1993session}, the assessment focuses on the following three aspects:
\textbf{(1) Session Outcome} considers the immediate impact of the counseling session, such as the client's progress toward their therapeutic goals and the perceived effectiveness of the sessions. 
\textbf{(2) Therapeutic Alliance} evaluates the quality of the relationship between the therapist and the client, emphasizing the client's trust in and agreement with the therapist. 
\textbf{(3) Self-Reported Feelings} capture the client's immediate emotions and provide a subjective view of their feelings about themselves and the session.
Four emotion dimensions are considered: \textit{Depth}, \textit{Positivity}, \textit{Smoothness}, and \textit{Arousal}.
All, particularly depth, relate to perceived session helpfulness \cite{cummings1995session,barak2006factors}, and arousal highly correlates with clients' ambivalence to change \cite{schneider2016let,engle2008viewing}.
See \Cref{sec:appendix:feelings} for more detailed information.
Correspondingly, we utilize five questionnaires commonly used in clinical psychology: SRS, CECS, SEQ, WAI-SR, and HAQ-II.

\subsubsection{Completion \& Assessment Results}
We prompt the LLM to complete the questionnaires.
The prompt includes the problems \& reasons for visiting, apparent traits, the interaction with the LLM therapist, as well as questionnaire items.
For each item, the rating scale and the scale meanings are provided in the prompt. The assessment result for each aspect is calculated using the relevant item scores. Higher values reflect better performance.

\section{Reliability of ClientCAST}
We examine the reliability of ClientCAST from two aspects: (1) the performances of simulated clients in counseling sessions and (2) the abilities of the assessment, based on questionnaire completion, to distinguish between high- and low-quality sessions.
\subsection{Performances of Simulated Clients}\label{sec:valid:client_simulation}
The simulated clients are expected to behave consistently with their psychological profiles and emulate real clients in counseling sessions.
We analyze the simulated clients' behaviors by involving the simulated client with the psychological profile $\mathcal{P}_{C}(S_i)$ in a new session $S'_i$.
We design two modes to obtain the simulated session $S'_i$: \textbf{(1) Simulated Client \text{\texttimes} LLM.} 
We use another LLM to simulate the therapist $T_i$ in the session $S_i$ and interact with the simulated client.
Specifically, we prompt the LLM to mirror $T_i$'s communication style and therapy skills, providing a reference session that rephrases $S_i$ to prevent a complete replication of $S_i$. 
All sessions and clients in datasets are simulated. 
\Cref{tab:sessions} displays an example of the simulated session. An additional example with the client's psychological profile is provided in \Cref{tab:appendix:sessions}.
\textbf{(2) Simulated Client \text{\texttimes} Human.} We recruit humans to interact with the simulated clients by mimicking the therapists' skills and speech tone. 
Ten sessions, randomly selected from the datasets, along with the corresponding clients, are simulated in this mode.
\setlength{\fboxsep}{1pt}
\begin{table}[ht!]
    \centering
    \scriptsize
    \begin{tabular}{m{.95\hsize}}
    \hline
    \colorbox[HTML]{6C8EBF}{\color{white}{\textbf{Therapist:}}} That's okay. It can be a bit unclear. What do you think led you to make an appointment with me?\\
    
    \colorbox[HTML]{82B366}{\color{white}{\textbf{Client:}}} I got caught taking money from my employer, so I had to come.\\
    
    \colorbox[HTML]{6C8EBF}{\color{white}{\textbf{Therapist:}}} So, getting caught taking money has led you to this point. How do you feel about what happened?\\
    
    \colorbox[HTML]{82B366}{\color{white}{\textbf{Client:}}} I didn't expect it to add up to what it did.\\
    
    \colorbox[HTML]{6C8EBF}{\color{white}{\textbf{Therapist:}}} Yeah, it's easy to get caught up in the moment and lose track. Can you tell me more about what was going through your mind when you took the money?\\
    
    \colorbox[HTML]{82B366}{\color{white}{\textbf{Client:}}} I don't know, it just happened.\\
    
    \colorbox[HTML]{6C8EBF}{\color{white}{\textbf{Therapist:}}} Okay. Sometimes, we do things without fully thinking them through. Did you feel like you were under any kind of pressure or stress at the time that might've contributed to your decision?\\
    
    \colorbox[HTML]{82B366}{\color{white}{\textbf{Client:}}} Nah, not really. I just wanted some extra cash.\\
    
    \colorbox[HTML]{6C8EBF}{\color{white}{\textbf{Therapist:}}} I get it. Extra cash can be tempting. How did you plan on using that extra money? Was it for something specific or just general spending?\\
    
    \colorbox[HTML]{82B366}{\color{white}{\textbf{Client:}}} Pokies, I suppose.\\
    \hline
    \end{tabular}
    \caption{A piece of an interaction between Llama 3-70B simulated client and therapist.}
    \label{tab:sessions}
\end{table}

\subsubsection{Simulated Clients' Verbal Styles}
We present the averaged response lengths and session turns\footnote{Since the end of the session is jointly determined by the client and therapist, we consider it to be when the client starts repeating the same response, such as ``thank you.''} between the $S_i$ and $S'_i$.
Additionally, to compare the language styles between the simulated client and the human client, we compute the vocabulary overlap rate and Language Style Matching (LSM) between them.
The word overlap between client responses in $S_i$ and $S'_i$ is computed by:
\begin{equation}
    \frac{V_{C}(S_i)\cap V_{C}(S'_i)}{\min(|V_{C}(S_i)|,|V_{C}(S'_i)|)},
\end{equation}
where $V_{C}(\cdot)$ denotes the vocabulary the client used in the session.
The LSM scores are computed using \cite{boyd2022development}. 
To validate that simulated clients' language style depends more on the psychological profile and reference sessions than the underlying models, we compute LSM between two different clients simulated by the same model.
Results are demonstrated in \Cref{tab:statistics_client_simulation}.

\begin{table}[ht!]
\centering
    \scriptsize
    \begin{tabular}{c|c|c|c|c}
    \hline
    \rowcolor{gray!20} Model & \makecell[c]{avg.\#len.\\response} & \makecell[c]{avg.\#ses\\sion turn} & \makecell[c]{avg.vocab\\overlap (\%)} & LSM\\ 
    \hline
    \multicolumn{4}{c}{\textit{(a) Simulated Client \text{\texttimes} LLM Therapist}}\\
    \hline
    Claude-3 & 72.84 & 66.52 & 38.83\% & \underline{\textbf{0.89}/0.88}\\
    GPT-3.5 & 7.51 & 43.91  & 29.72\% & \underline{0.80/0.74}\\
    Llama 3-70B & \textbf{11.87} & \textbf{54.77} & \textbf{45.44\%} & \underline{\textbf{0.89}/0.82}\\
    Mixtral 8\text{\texttimes}7B & 37.80 & 63.12 & 29.22\% & 0.81/0.85\\
    \hline
    Human & 22.61 & 53.87 & - & - /0.85\\
    \hline
    \multicolumn{4}{c}{\textit{(b) Simulated Client \text{\texttimes} Human Therapist}}\\
    \hline
    Claude-3 & 51.17 & 35.60 & 22.78\% & \underline{\textbf{1.00}/0.86}\\
    GPT-3.5 & \textbf{10.44} & 28.60  & 38.50\% & \underline{0.90/0.83}\\
    Llama 3-70B & 18.81 & \textbf{36.60} & \textbf{56.66\%} & \underline{0.92/0.85}\\
    Mixtral 8\text{\texttimes}7B & 47.42 & 52.60 & 25.66\% & 0.93/0.91\\
    \hline
    Human & 12.80 & 41.40 & - & - /0.86 \\
    \hline
    \end{tabular}
\caption{Statistics of simulated clients' verbal styles. 
\textbf{Bold values} indicate the \textbf{most similarity to human clients}.
For LSM, the two values represent LSM between (the simulated and human clients)/(two clients simulated by the same LLM). A larger LSM value indicates more similar language styles between the two clients. The \underline{underlined LSM} pairs indicate that the former LSM is \underline{statically significantly larger} than the latter with p-value$<$0.05. According to the official LIWC-22 tutorials \cite{Pennebaker2023}, the acceptable LSM value in conversations generally ranges from 0.83 to 0.94.}
\label{tab:statistics_client_simulation}
\end{table}

From the results, \textbf{(1) Clients simulated by different LLMs demonstrate various language styles.} For example, Claude-3 tends to generate longer responses, whereas GPT-3.5 tends to produce shorter ones.
\textbf{(2) Simulated clients can mimic the human clients' language styles}, except for clients simulated by Mixtral 8\text{\texttimes}7B.
It can be observed that the LSM between the simulated and human clients is usually statistically significantly higher than the LSM between two different clients simulated by the same LLM.
This demonstrates that simulated clients' language styles depend more on the provided psychological profiles and reference sessions than on their underlying LLMs.

\subsubsection{Client Consistency Results \& Analysis}
\begin{table*}[tb]
    \centering
    \scriptsize
    \begin{tabular}{c|ccc|cc|cc}
    \hline
    \rowcolor{gray!20}  & \multicolumn{3}{c|}{Problems \& Reasons for Visiting} & \multicolumn{2}{c|}{Symptoms} & \multicolumn{2}{c}{Apparent Traits} \\
    \hhline{>{\arrayrulecolor{gray!20}}-|>{\arrayrulecolor{black}}-------}
    \rowcolor{gray!20}\multirow{-2}{*}{Model} & Problems Similarity & Reason Similarity & Session Topic Precision & Recall & F1 & Recall & F1 \\
    \hline
    \multicolumn{8}{c}{\textit{(a) Simulated Client \text{\texttimes} LLM Therapist \& Automatic Evaluation}}\\
    \hline
    Claude-3 & \textbf{73.02\%} \scriptsize{(0.72/0.19)} & \textbf{70.98\%} \scriptsize{(0.74/0.22)}  & 0.90 & 0.77 & 0.69 & 0.71 & 0.72\\
    
    GPT-3.5 &  64.25\% \scriptsize{(0.64/0.23)} & 65.57\% \scriptsize{(0.67/0.23)} & 0.92 & 0.84 & \textbf{0.90} & 0.59 & 0.60  \\
    
    Llama 3-70B & 72.71\% \scriptsize{(0.72/0.20)} & 68.02\% \scriptsize{(0.77/0.25)} & \textbf{0.94} & \textbf{0.86} & 0.85 & \textbf{0.78} & \textbf{0.78} \\
    
    Mixtral 8\text{\texttimes}7B  & 58.32\% \scriptsize{(0.74/0.31)} & 33.84\% \scriptsize{(0.81/0.54)} & 0.85 & 0.68 & 0.74 & 0.58 & 0.57 \\
    \hline
    \multicolumn{8}{c}{\textit{(b) Simulated Client \text{\texttimes} Human Therapist \& Human Rating}}\\
    \hline
    Claude-3 & - & - & \textbf{1.00} & 0.93 & 0.78 & \textbf{0.84} & \textbf{0.83} \\
    GPT-3.5 &  - & - & \textbf{1.00} & \textbf{0.98} & \textbf{0.98} & 0.62 & 0.67 \\
    Llama 3-70B & - & - & \textbf{1.00} & \textbf{0.98} & 0.95 & 0.82 & 0.81\\
    Mixtral 8\text{\texttimes}7B  & - & - & \textbf{1.00} & 0.93 & 0.86 & 0.79 & 0.83\\
    \hline
    \end{tabular}
    \caption{Results when comparing the original psychological profile and the one extracted the simulated session. A larger value indicate the simulated client's higher consistency with the original psychological profile. For sentence similarity, we provide the absolute similarity values of (\textit{the target pairs} \textbf{/} \textit{the randomly selected pairs}) as a reference.}
    \label{tab:client_simulation}
\end{table*}

% Claude-3 & - & - & 1.00 & 1.00 & 0.81 & 0.77 & 0.82 & 0.81  \\
% GPT-3.5 &  - & - & 0.80 & 0.72 & 0.90 & 0.92 & 0.60 & 0.66  \\
% Llama 3-70B & - & - & 0.80 & 0.73 & 0.97 & 0.88 & 0.72 & 0.72 \\
% Mixtral 8\text{\texttimes}7B  & - & - & 0.80 & 0.87 & 0.86 & 0.88 & 0.64 & 0.68 \\
To examine the extent to which the simulated client behaves consistently with the given psychological profile, we compare the psychological profile $\mathcal{P}_{C}(S'_i)$ extracted from $S'_i$ with the original one $\mathcal{P}_{C}(S_i)$. 
The comparison regarding problems \& reasons for visiting is based on sentence similarity.
However, we do not directly present the absolute similarity values, considering the inherent similarities among texts in the same domain and those generated by the same model.
Thus, we measure a \textit{normalized relative similarity (\%)}:
\begin{equation}
1-\frac{\text{similarity}_\text{(random pairs)}}{\text{similarity}_\text{(target pairs)}},
\end{equation}
where the target pairs are the problems \& reasons for visiting sentences of the human client and its simulated one, and the random pairs are problems \& reasons for visiting of two different clients simulated by the same model.
In addition, we also present the precision of the session topic of the new session $S'_i$. We use seven topics covered by the datasets.
For the comparison of symptoms and apparent traits in $\mathcal{P}_{C}(S_i)$ and $\mathcal{P}_{C}(S'_i)$, we employed \textit{recall} and \textit{F1 score}.
For sessions simulated by Simulated Clients \text{\texttimes} Human, we ask human annotators to extract the psychological profiles from these sessions and conduct comparisons.
\Cref{tab:client_simulation} presents the results.
Values in subtables (a) and (b) are averages over 300 and 10 sessions, respectively.

It can be observed: \textbf{(1) The performances of simulated clients are determined significantly by the underlying LLM.} Stronger LLMs tend to achieve higher scores. Clients simulated by Claude-3 and Llama 3-70B perform better, as they achieve higher scores across most metrics. 
\textbf{(2) Simulated clients perform better at presenting problems \& reasons for visiting and symptoms, but they are less effective at displaying the assigned apparent traits.}
The metric values for problems \& reasons for visiting are generally high, except for clients simulated by Mixtral 8\text{\texttimes}7B, which consistently achieve the worst performance.
The session topic precision is not 1 because a single session can sometimes encompass multiple topics, leading the simulated session to focus on a different topic than the original one. 
Since the therapist also determines the main session topic, this discrepancy cannot be attributed to the inconsistency of the simulated client.
Moreover, the recall and F1 scores for symptoms across 61 labels are relatively high.
However, the scores for apparent traits across 3 to 5 classes are comparatively low.
One possible reason can be that LLM therapists do not perform as human therapists, while the apparent traits are easily influenced by the behavior of the other interlocutor \cite{zhang2022real}.
This is evident from the fact that, for the \textit{same} 10 simulated clients in the Simulated Client\text{\texttimes} LLM interactions, the recall and F1 scores decreased by 1.2\%$\sim$ 27\% compared to the values in \Cref{tab:client_simulation} (b).
Furthermore, as depicted in \Cref{tab:appendix:traits} in \Cref{sec:appendix:client_simulation}, apparent traits such as openness, emotional fluctuations, and resistance toward the therapist in new sessions are less consistent with the psychological profiles than other apparent traits.
Surprisingly, the neuroticism of the simulated clients, which encompasses more complex characteristics, can be fairly consistent. 
This is likely because symptoms in the psychological profile can reflect the neuroticism of a simulated client to some extent, allowing this trait to be accurately represented in new sessions. 

\begin{figure}[htb]
\centering
\includegraphics[width=\linewidth]{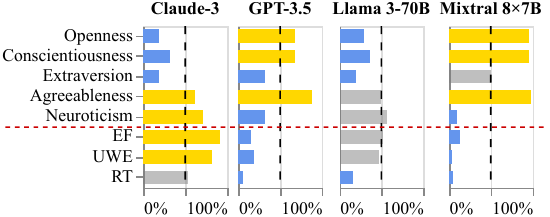} 
\caption{The proportion of inconsistent simulated clients who exhibit a higher level of apparent traits. EF: Emotion Fluctuations, UWE: UnWillingness to express emotions, RT: Resistance toward the Therapist.}
\label{fig:personality_error}
\end{figure}

By further error analysis, we have the following findings:
\textbf{(1) Clients simulated by GPT-3.5 are more resilient, while those simulated by Llama 3-70B are more unenthusiastic.}
It has been argued that different LLMs exhibit bias towards different personalities \cite{jiang2023personallm}.
We analyze the inconsistent simulated clients, i.e., the apparent traits observed in $S'_i$ are different from those observed in $S_i$, and explore the influence of LLMs on the apparent traits.
Specifically, we compute the proportion of the inconsistent simulated clients who perform a higher level of the apparent traits in $S'_i$ than in $S_i$, demonstrated in \Cref{fig:personality_error}.
The proportion is expected to be 50\% if the LLMs do not exhibit a tendency towards specific traits. 
It can be inferred that clients simulated by Mixtral 8\text{\texttimes}7B and GPT-3.5 tend to be more resilient, whereas clients simulated by Claude-3 and Llama 3-70B tend to be more sensitive.
Moreover, Llama 3-70B simulated clients appear less enthusiastic.
The biases in different models can be leveraged to simulate various clients.
\textbf{(2) Symptoms with general descriptions are easier to simulate.}
Through error analysis, it has been observed that symptoms described in general terms, such as ``\textit{feeling nervous, anxious, or on edge},'' are easier to simulate than those with more specific descriptions, such as ``\textit{feeling afraid of open spaces, of driving, or being on buses, subways, and so forth}.'' 
Notably, these generally described symptoms can be performed by the simulated clients even when they are not included in the psychological profile.
Fortunately, these additional symptoms do not lead to client inconsistency, although they may influence the symptom F1 scores in \Cref{tab:client_simulation}.
Further, the effectiveness of symptom simulation does not clearly correlate with the symptom categories. 
For instance, the rates of success or failure in simulating symptoms related to social roles and depression are comparable.

\subsection{Efficacy of Assessment by Questionnaires}
\begin{figure*}[ht!]
\centering
\includegraphics[width=\linewidth]{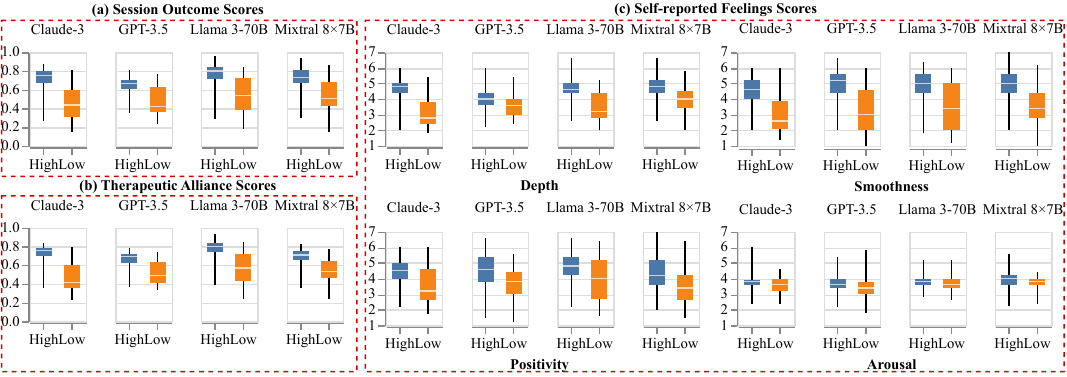} 
\caption{Session outcome, therapeutic alliance and self-reported feelings scores of high- and low-quality sessions in High-Low Quality Counseling and AnnoMI datasets.}
\label{fig:questionnaire_results}
\end{figure*}
For each session $S_i$ in the datasets, we ask the simulated client with psychological profile $\mathcal{P}_{C}(S_i)$ to complete questionnaires based on $S_i$. 
Then, we compare the assessment of high- and low-quality sessions according to the completed questionnaires. 

\subsubsection{Assessment Results on Datasets}
Comparison results are presented in \Cref{fig:questionnaire_results}.
The findings are as follows: \textbf{(1) High- and low-quality sessions can be distinguished clearly in terms of the session outcomes and therapeutic alliance}, especially when the underlying model is Claude-3 or Llama 3-70B.
However, there can be several outliers where clients exhibit a severe lack of motivation to address their problems or require immediate solutions from the therapists. 
Consequently, it is inherently challenging for therapists to engage these clients in counseling sessions \cite{swift2011decreasing,bados2007efficacy}, which can result in low scores.
\textbf{(2) Positivity and arousal scores do not distinguish between high- and low-quality sessions.}
However, this phenomenon is reasonable and reflects the nature of therapy.
Arousal is related to clients' ambivalence to change \cite{schneider2016let,engle2008viewing}, which primarily manifests at the end of long-term therapy, spanning months. 
However, most sessions in our datasets are at the early to middle stages of therapy. 
Therefore, the arousal scores, even for high-quality sessions, are not high.
Positivity and smoothness, particularly the former, can be influenced by the therapist's strategy. For example, in an exploratory session, clients usually exhibit less smoothness and positivity compared to a perspective session \cite{mallinckrodt1993session}.
Nevertheless, low scores regarding positivity and smoothness in high-quality sessions are less likely to be related to dissatisfaction with therapists or sessions compared to low scores in low-quality sessions.
To prove this, we conduct LIWC analysis of LLM-generated explanations for the self-reported feelings, shown in \Cref{tab:liwc_explanations}, and have the additional finding:
\textbf{(3) Explanations for the self-reported feelings significantly differ in high- and low-quality counseling sessions.} 
The simulated clients in high-quality sessions use a more positive tone when describing their feelings, and their negative emotions are mainly characterized by anxiety related to their problems. 
Conversely, clients in low-quality sessions exhibit more anger in their negative emotions, and they express dissatisfaction with therapists or sessions much more frequently.
% than clients in high-quality sessions.

\begin{table}[ht!]
    \scriptsize
    \centering
    \begin{tabular}{l|cccc}
        \hline
        \multirow{2}{*}{\makecell[l]{Word\\Attribute}} & Claude-3 & GPT-3.5& \makecell[c]{Llama 3\\-70B}& \makecell[c]{Mixtral\\8\text{\texttimes}7B}\\
        & High\hspace{.1cm}Low & High\hspace{.1cm}Low & High\hspace{.1cm}Low & High\hspace{.1cm}Low \\
        \hline
        Pos Tone & 5.62\hspace{.15cm}3.34 & 5.52\hspace{.15cm}3.84 & 5.38\hspace{.15cm}3.47 & 4.81\hspace{.15cm}2.98 \\
        Neg Tone & 3.84\hspace{.15cm}4.93 & 3.77\hspace{.15cm}5.86 & 3.66\hspace{.15cm}5.76 & 3.12\hspace{.15cm}4.41 \\
        Pos Emotion & 1.58\hspace{.15cm}0.99 & 2.04\hspace{.15cm}1.51 & 2.25\hspace{.15cm}1.61 & 1.83\hspace{.15cm}1.08 \\
        Neg Emotion & 2.85\hspace{.15cm}3.30 & 2.60\hspace{.15cm}4.01 & 2.84\hspace{.15cm}4.81 & 2.56\hspace{.15cm}3.41 \\
        Anxiety & 1.76\hspace{.15cm}1.10 & 1.67\hspace{.15cm}1.77 & 1.83\hspace{.15cm}1.52 & 1.82\hspace{.15cm}2.12 \\
        Anger & 0.24\hspace{.15cm}0.67 & 0.24\hspace{.15cm}0.68 & 0.41\hspace{.15cm}2.10 & 0.14\hspace{.15cm}0.33 \\
        Sadness & \textit{0.28}\hspace{.15cm}\textit{0.37} & 0.21\hspace{.15cm}0.34 & 0.13\hspace{.15cm}0.27 & \textit{0.20}\hspace{.15cm}\textit{0.28} \\
        \hline
    \end{tabular}
    \caption{LIWC analysis of explanations for self-reported feelings. Except for the \textit{values in italics}, the values for high- and low-quality sessions differ statistically significantly, with p-value$<$0.01. A higher value indicates a more severe level of clients' tones/emotions.}
    \label{tab:liwc_explanations}
\end{table}

Considering the above results and analyses, we have decided to \textbf{use GPT-3.5 and Llama 3-70B to simulate resilient and sensitive clients}, respectively, within the ClientCAST framework.
\section{Evaluation of LLM Therapists}\label{sec:evaluation}
\subsection{LLM Therapists}
A common way current LLMs are being used as therapists is through custom ``system prompts'' that instruct them to function as therapists. We use a simple ``system prompt'', shown in \Cref{fig:appendix:therapist_prompt}, which is designed based on prompts used in \cite{chiu2024computational} and \cite{chen2023llm}. 

\begin{figure*}[ht!]
\centering
\includegraphics[width=\linewidth]{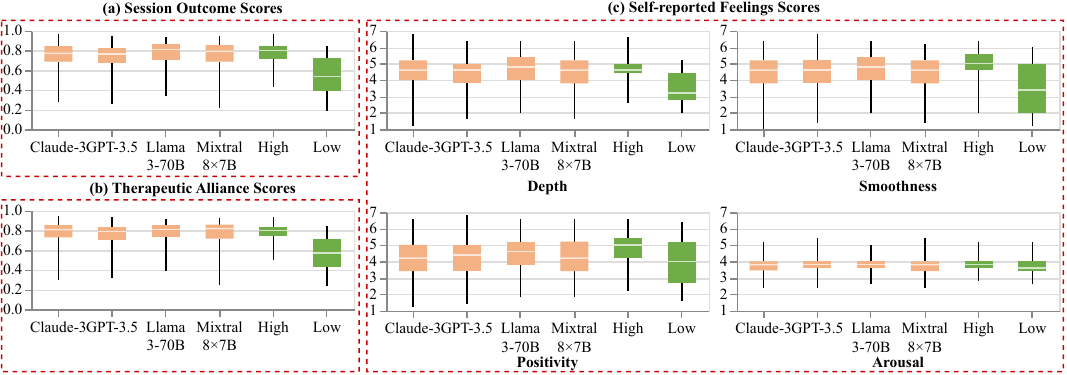} 
\caption{LLM therapist assessments on session outcome, therapeutic alliance, and self-reported feelings using ClientCAST.}
\label{fig:therapist_assessment}
\end{figure*}
\subsection{Client-centered Assessment}
We adopt ClientCAST to assess LLM therapists, as illustrated in \Cref{fig:clientcast}.
The results are shown in \Cref{fig:therapist_assessment}. 
For reference, the assessment includes both high- and low-quality human-human counseling sessions.
We conduct significance tests between the assessment scores of sessions involving simulated clients with psychological profiles extracted from high- and low-quality sessions. 
The resulting p-values are all $1.00$, indicating that the source of the client's psychological profiles does not influence the assessment.
Here are the findings from the results: 
\textbf{(1)} The performance of LLM therapists is significantly influenced by the underlying LLM. 
Generally, more powerful LLMs achieve higher and more stable scores.
\textbf{(2)} LLM therapists can foster strong connections with clients. They achieve comparable scores in terms of therapeutic alliance. Additionally, their session outcome scores are high but slightly lower than those of human therapists in high-quality sessions.
\textbf{(3)} LLM therapists are disadvantaged in reacting to clients' emotions.
The self-reported feelings scores, especially regarding positivity and smoothness, are lower than those of human therapists in high-quality sessions despite being higher than those in low-quality sessions.
In addition to the inherent difficulty of always maintaining high positivity and smoothness during effective therapy \cite{fullerton2021integrative,mallinckrodt1993session}, LLM therapists also struggle to react effectively to clients' emotions.
LIWC analysis (\Cref{tab:appendix:liwc_therapist}) shows that LLMs' responses focus significantly more on emotions, particularly positive ones, compared to human therapists. 
In LLM therapist responses, the frequency of emotion-related words is 1.27$\sim$2.16 times that in human therapist responses. 
Notably, LLM therapists may inappropriately and excessively employ responses such as ``It is understandable...'' 
or ``That is really tough...''
Instead, human therapists, especially in high-quality sessions, prefer helping clients to explore their negative feelings, such as anxiety; they tend to use terms such as 'what' and 'if' more frequently.

\subsection{LLM Therapists' Verbal Styles}
\begin{table}[ht!]
\centering
    \scriptsize
    \begin{tabular}{c|c|c|c|c}
    \hline
    Model & \makecell[c]{avg.\#len.\\response} & \makecell[c]{avg.\#ses\\-sion turn} & \makecell[c]{avg.vocab\\overlap} & \makecell[c]{LSM\\high/low/self}\\ 
    \hline
    Claude-3 & 83.03 & 61.91 & 21.60\% & \textbf{0.85}/0.83/0.88\\
    GPT-3.5 & 58.38 & 61.25  & 22.27\%& 0.80/0.78/0.83\\
    \makecell{Llama 3-70B} & \textbf{48.73} & 62.73 & \textbf{23.34\%}& \textbf{0.85}/0.81/0.84\\
    \makecell[c]{Mixtral 8\text{\texttimes}7B} & 71.90 & \textbf{62.76} & 20.46\%& 0.81/0.79/0.88\\
    \hline
    High & 25.15 & 65.42 & - & - /0.85/0.87\\
    Low & 28.25 & 38.43 & - & - /0.85/0.84\\
    \hline
    \end{tabular}
\caption{Statistics of LLM therapists' verbal style. High/Low represent human therapists in the high/low-quality sessions. \textbf{Bold values} indicate \textbf{most similarity to human therapists in high-quality sessions}.}
\label{tab:statistics_therapist}
\end{table}
\Cref{tab:statistics_therapist} presents LLM therapists' verbal styles.
The analysis includes the LSM between LLM therapists and human therapists in high- and low-quality sessions, as well as the LSM between therapists implemented by the same LLM. 
Compared to human therapists, LLM therapists tend to generate longer utterances. 
The language styles of different therapists implemented by the same LLM are more similar to each other than to those of human therapists. 
This is likely because we did not instruct the LLMs to mimic the styles of human therapists. 
\section{Conclusion and Discussion}
This work proposes a client-centered approach to assessing LLM therapists.
We involve clients in LLM therapist assessment by leveraging LLMs to simulate clients.
Simulated clients are used to interact with LLM therapists and complete questionnaires about the interaction. 
Then, the client-centered assessment results are derived from the completed questionnaires. 
Through experiments, we find that LLMs can generally, though not perfectly, simulate clients, and they are able to distinguish high- and low-quality sessions by completing client-centered questionnaires. 
Then, we assess various LLM therapists using ClientCAST.

We should acknowledge that, in the short term, LLMs struggle to achieve perfect simulation and high levels of human trust. 
However, as argued by \citet{yang2024social}, the imperfect simulation of LLMs can benefit humans in exploring specific tasks. 
Furthermore, these imperfections can be viewed as a trade-off to avoid the high costs associated with involving humans in experiments.
In our task, we use LLM-simulated clients to assess LLM therapists. 
This approach is not based on the belief that LLMs are better at simulating clients than therapists. 
Instead, we believe the simulated clients can provide an environment that allows LLMs to demonstrate their capabilities and limitations as therapists. 
The client-centered assessment also offers a direction for further analyses and studies.

% \newpage
\section*{Limitations}
One limitation in the simulation of human behavior is inconsistency, which we also observe in the field of counseling therapy. 
Neither the simulation of therapists nor clients is perfect. 
In this work, LLMs face challenges in accurately simulating the personalities of human clients. 
However, our experiments show that more powerful LLMs achieve higher simulation consistency and accuracy. 
Additionally, different LLMs exhibit inconsistency in various ways. 
Therefore, on the one hand, we believe that more advanced LLMs can mitigate this issue in the future. 
On the other hand, as practiced in this work, we can leverage the inherent biases of current LLMs to simulate characters, such as clients, with diverse features.

\section*{Ethical Considerations}
\paragraph{Social Impact}
This work does not advocate for the use of LLMs in therapy. 
Instead, we propose an assessment approach to reveal the characteristics of LLM therapists for further study, particularly given the surge of users who have been using LLMs as therapists.
We do not suggest that LLMs can replace human workload. 
As mentioned, such an assessment approach can help reduce the cost of human-simulated clients and mitigate some associated risks. 
Certainly, LLMs' outputs are not as expert and accurate as those of professionals, but they can still be used as supplementary tools and inspire human exploration.
Therefore, we hope the ideas presented in this work provide NLP and psychology researchers with an alternative method for applying LLMs in the automatic evaluation of counseling sessions, fostering further discussions on this topic. This can facilitate future research in AI psychology and sociology.
\paragraph{Human Annotations}For the human annotations, we emphasized the comfort and well-being of our annotators. The human annotators have been well paid for their efforts.
\paragraph{Use of Datasets}In our experiments, we adopted open-sourced datasets compliance with the appropriate licenses and consents the authors provide within their terms of use.

Consequently, we confidently assert that our research is conducted in strict adherence to the ethical guidelines prescribed by the Association for Computational Linguistics.

\bibliography{custom}

\appendix
\section{Client Simulation}
\subsection{Psychological Profile}\label{sec:appendix:client_information}
\begin{table}[ht!]
    \centering
    \scriptsize
    \begin{tabular}{m{.95\hsize}}
    \hline
    \textbf{Name:} Ricky \\
    \textbf{Gender:} Male \\
    \textbf{Age:} Late 20s. This estimate is based on Ricky's concern about his acting career and his social life, which suggests that he is old enough to have established a career and social connections but still young enough to be figuring out his priorities and struggling with drug use.\\
    \textbf{Occupation:} Actor.\\
    \textbf{Problem:} Substance abuse. Ricky is struggling with drug use, specifically meth, which is affecting his life, including his acting career, social relationships, and overall well-being.\\
    \textbf{Reasons for Visiting:} Ricky is visiting the therapist because his friends are worried about his drug use, specifically meth, and how it's affecting his acting career and overall well-being.\\
    \textbf{Apparent Traits:} \\
    \underline{Openness} is approximately 0-20\%. Ricky appears to be stuck in his current situation and is not exhibiting a strong desire to explore new ideas, experiences, or perspectives, instead relying on familiar habits and social connections.\\
    \underline{Conscientiousness} is approximately 0-20\%. Ricky's lack of motivation, emphasis on short-term pleasure, and tendency to prioritize social life over personal goals and responsibilities suggest a relatively low level of conscientiousness.
    \underline{Extraversion} is approximately 60-80\%, a moderate to high level of extraversion, as Ricky frequently mentions his social life, friends, and desire to fit in, and can appear to prioritize social connections and having fun over his acting career.\\
    \underline{Agreeableness} is approximately 60-80\%. Ricky exhibits some cooperative and friendly traits, such as acknowledging his friends' concerns and showing appreciation for the therapist's help, but also displays some ambivalence and resistance to change, indicating a moderate level of agreeableness.\\
    \underline{Neuroticism} is approximately 60-80\%. Ricky exhibits some traits of neuroticism, such as feeling anxious, uncertain, and judgmental about himself and his situation, but also shows a willingness to acknowledge his fears and uncertainties, and a desire to make changes, indicating some emotional stability and hope for improvement.\\
    \underline{Emotion Fluctuation} is Medium. Ricky's emotions fluctuate moderately, as he oscillates between feelings of frustration, annoyance, and uncertainty, but also experiences moments of hope, relief, and motivation throughout the conversation.\\
    \underline{Unwillingness to Express Feelings} is Low. Ricky is relatively willing to express his feelings, as he openly discusses his concerns, emotions, and desires throughout the conversation.\\
    \underline{Resistance towards the Therapist} is Low. Ricky exhibits a low level of resistance towards the therapist, as he is willing to engage in the conversation, shares his concerns and feelings, and shows appreciation for the therapist's help.\\
    \textbf{Symptoms:}\\
    \underline{Feeling down, depressed, or hopeless}: While Ricky doesn't explicitly express feelings of depression or hopelessness, Ricky can mention feeling stuck, unmotivated, and disconnected from his passion for acting. He also expresses a desire to feel like himself again, which suggests a sense of dissatisfaction with his current emotional state.\\
    \underline{Feeling bad about myself or that I am a failure or have let myself or my} \underline{family down}: Ricky has expressed feelings of being stuck and wanting to get his life back on track, indicating a sense of disappointment or dissatisfaction with his current situation. \\
    \underline{Having trouble at work/school because of drinking or drug use}: He mentions missing a couple of auditions and not getting as many jobs as he used to, which indicates that his drug use is affecting his acting career, but it's not a consistent or frequent issue.\\
    \hline
    \end{tabular}
    \caption{An example of the psychological profile.}
    \label{tab:appendix:profile}
\end{table}

We provide the example of the psychological profile in \Cref{tab:appendix:profile}.
We account for 61 client symptoms in the following three questionnaires, which are widely used in clinical psychology:
\paragraph{Patient Health Questionnaire (PHQ-9)}
The nine-item depression module from the PHQ-9 is well validated and widely used as a brief diagnostic and severity measure \cite{kroenke2001phq}. \Cref{tab:appendix:PHQ-9} illustrate the content of PHQ-9.
\begin{table}[ht!]
   \scriptsize
    \centering
    \begin{tabular}{m{.03\hsize}m{.85\hsize}}
    \hline
    \textbf{No} & \textbf{PHQ-9 Item} \\
    \hline
    1 & Little interest or pleasure in doing things. \\
    2 & Feeling down, depressed, or hopeless. \\
    3 & Trouble falling or staying asleep, or sleeping too much. \\
    4 & Feeling tired or having little energy. \\
    5 & Poor appetite or overeating. \\
    6 & Feeling bad about yourself or that you are a failure or have let yourself or your family down. \\
    7 & Trouble concentrating on things, such as reading the newspaper or watching television. \\
    8 & Moving or speaking so slowly that other people could have noticed. Or the opposite being so figety or restless that you have been moving around a lot more than usual. \\
    9 & Thoughts that you would be better off dead, or of hurting yourself. \\
    \hline
    \end{tabular}
    \caption{PHQ-9 items. Each question starts with ``\textit{Over the last 2 weeks, how often have you been bothered by any of the following problems?}''}
    \label{tab:appendix:PHQ-9}
\end{table}

\paragraph{Generalized Anxiety Disorder 7-item Scale (GAD-7)} 
The GAD-7 \cite{spitzer2006brief} is a brief self-report scale designed to identify probable cases of generalized anxiety disorder. It is widely used for adults in many different cultures \cite{seo2015validation}. The GAD-7 is similar to the PHQ-9 but with slight differences in their focus on the client's symptoms
\cite{pranckeviciene2022validation}. \Cref{tab:appendix:GAD-7} shows the content of GAD-7.
\begin{table}[ht!]
    \scriptsize
    \centering
    \begin{tabular}{ll}
    \hline
    \textbf{No} & \textbf{GAD-7 Item} \\
    \hline
    1 & Feeling nervous, anxious, or on edge. \\
    2 & Not being able to stop or control worrying. \\
    3 & Worrying too much about different things. \\
    4 & Trouble relaxing. \\
    5 & Being so restless that it is hard to sit still. \\
    6 & Becoming easily annoyed or irritable. \\
    7 & Feeling afraid as if something awful might happen. \\
    \hline
    \end{tabular}
    \caption{GAD-7 items. Each question starts with ``\textit{Over the last two weeks, how often have you been bothered by any of the following problems?}''}
    \label{tab:appendix:GAD-7}
\end{table}

\paragraph{Outcome Questionnaire-45 (OQ-45)}
The OQ-45 was designed to measure client progress in therapy and to be repeatedly administered during treatment and at termination \cite{lambert1996administration,lambert2004outcome}. However, we adopt it to assess the client's symptoms because it assesses three aspects of the client's life, i.e., subjective discomfort, problems in interpersonal relationships, and problems in social role performance. The items also measure personally and socially relevant characteristics that affect the individual's quality of life, attempting to quantify both positive and negative functioning. OQ-45 items are shown in \Cref{tab:appendix:oq-45}.

\begin{table}[ht!]
    \scriptsize
    \centering
    \begin{tabular}{m{.02\hsize}m{.37\hsize}|m{.02\hsize}m{.37\hsize}}
    \hline
    \textbf{No} & \textbf{OQ-45 Item} &\textbf{No} & \textbf{OQ-45 Item} \\
    \hline
    1 & I get along well with others. & 24 & I like myself. \\
    2 & I tire quickly. & 25 & Disturbing thoughts come into my mind that I cannot get rid of. \\
    3 & I feel no interest in things. & 26 & I feel annoyed by people who criticize my drinking (or drug use). (If not applicable, mark ``never'') \\
    4 & I feel stressed at work/school. & 27 & I have an upset stomach. \\
    5 & I blame myself for things. & 28 & I am not working/studying as well as I used to. \\
    6 & I feel irritated. & 29 & My heart pounds too much. \\
    7 & I feel unhappy in my marriage/significant relationship. & 30 & I have trouble getting along with friends and close acquaintances. \\
    8 & I have thoughts of ending my life. &  31 & I am satisfied with my life. \\
    9 & I feel weak. & 32 & I have trouble at work/school because of drinking or drug use. (If not applicable, mark ``never'') \\
    10 & I feel fearful. & 33 & I feel that something bad is going to happen. \\
    11 & After heavy drinking, I need a drink the next morning to get going. (If you do not drink, make ``never'') & 34 & I have sore muscles. \\
    12 & I find my work/school satisfying. & 35 & I feel afraid of open spaces, of driving, or being on buses, subways, and so forth. \\
    13 & I am a happy person. & 36 & I feel nervous. \\
    14 & I work/study too much. & 37 & I feel my love relationships are full and complete. \\
    15 & I feel worthless. & 38 & I feel that I am not doing well at work/school. \\
    16 & I am concerned about family troubles. & 39 & I have too many disagreements at work/school. \\
    17 & I have an unfulfilling sex life. & 40 & I feel something is wrong with my mind. \\
    18 & I feel lonely. & 41 & I have trouble falling asleep or staying asleep. \\
    19 & I have frequent arguments. & 42 & I feel blue. \\
    20 & I feel loved and wanted. & 43 & I am satisfied with my relationships with others. \\
    21 & I enjoy my spare time. & 44 & I feel angry enough at work/school to do something I might regret. \\
    22 & I have difficulty concentrating. &  45 & I have headaches. \\
    23 & I feel hopeless about the future. & &\\
    \hline
    \end{tabular}
    \caption{OQ-45 items.}
    \label{tab:appendix:oq-45}
\end{table}

These symptoms encompass depression (assessed by PHQ-9), anxiety (assessed by GAD-7), and symptom distress (assessed by OQ-45 items 2, 3, 5, 6, 8-10, 13. 15. 22-25, 29. 31, 33-36, 41, 42 \& 45), interpersonal relations (assessed by OQ-45 items 4, 12, 14, 21, 28, 32, 38, 39 \& 44), and social roles (assessed by OQ-45 items 1, 7, 16-20, 26, 30, 37 \& 43).
There can be some confusion among depression, anxiety and symptom distress. 
\textit{Depression} is a mood disorder marked by continuous sadness, hopelessness, and loss of interest in activities, while \textit{anxiety} involves excessive worry and fear about future events, accompanied by physical symptoms like restlessness and tension. 
\textit{Distress}, on the other hand, is a general state of emotional suffering that can arise from various stressors and is not a clinical diagnosis. Depression and anxiety are diagnosable mental health conditions with specific criteria, whereas distress is a broader emotional response that can occur in reaction to life changes or stress. 
Each condition has unique symptoms and treatment approaches, with depression and anxiety often requiring psychotherapy and medications and distress being managed through stress relief and coping strategies.

\begin{table*}[ht!]
\centering
    \begin{minipage}[t]{.46\linewidth}
    \centering
    \tiny
    \begin{tabular}{|m{\linewidth}|}
    \hline
     \cellcolor{gray!20} \makecell[c]{\textbf{Client Psychological Profile}}\\
    \hline
    \textbf{Name:} Not specified.\\
    \textbf{Gender:} Male.\\
    \textbf{Age:} Late teens to early twenties; your language and concerns (e.g., community service, caseworker, summer break) suggest they are likely a young adult, possibly a college student or recent high school graduate.\\
    \textbf{Occupation:} Not Specified. \\
    \textbf{Problem:} \\
    Legal issue: Completing community service hours as a requirement.\\
    \textbf{Reasons for Visiting:} \\
    The client is visiting the therapist because they have community service requirements to fulfill and need guidance on completing them, and potentially for related personal or emotional support.\\
    \textbf{Apparent Traits:} \\
    \underline{Openness} is approximately 0-20\%. The client appears to be fairly rigid and stuck in his ways, showing limited curiosity or desire to explore new ideas or experiences, and instead seems to focus on getting by with the minimum effort required.\\
    \underline{Conscientiousness} is approximately 20-40\%. The client displays a lack of organization, planning, and self-discipline, often appearing uncertain and easily distracted, which suggests a relatively low level of conscientiousness.\\
    \underline{Extraversion} is approximately 0-20\%. The client appears to be reserved and lacks assertiveness, preferring to seek guidance and reassurance from the therapist rather than taking initiative or expressing strong emotions or opinions.\\
    \underline{Agreeableness} is approximately 40-60\%. The client demonstrates some cooperative and respectful traits, such as acknowledging the therapist's help and expressing gratitude, but also shows a lack of enthusiasm and initiative, and a tendency to be somewhat detached and unclear in his responses. \\
    \underline{Neuroticism} is approximately 20-40\%. The client exhibits some anxiety and uncertainty, but primarily focuses on finding solutions and making a plan, demonstrating a relatively low level of emotional reactivity and dissatisfaction with life.\\
    \underline{Emotion Fluctuation} is Low. The client's emotions seem to remain relatively stable, with no significant shifts in tone or emotional intensity throughout the conversation. \\
    \underline{Unwillingness to Expression Feelings} is Medium. The client seems hesitant to express feelings, but still shares some concerns and emotions, such as feeling worried and uncertain, indicating a moderate level of unwillingness to express feelings. \\
    \underline{Resistance towards the Therapist} is Low. The client appears cooperative and receptive to the therapist's suggestions and guidance throughout the conversation.\\
    \textbf{Symptoms:}\\
    \underline{Trouble concentrating on things, such as reading the newspaper or watching television}: The client suggests some difficulty paying attention to details, as evidenced by his half-listening to his caseworker and uncertainty about the community service process. This lack of attention to detail might be indicative of some trouble concentrating, but it's not a prominent feature of the conversation.\\
    \underline{Feeling nervous, anxious, or on edge}: The client exhibits some symptoms of feeling nervous or anxious, such as worrying about getting community service hours done on time, feeling uncertain about the process, and being concerned about remembering to follow through on his plan. However, these symptoms do not can appear to be intense or pervasive, and he is able to lightly acknowledge and discuss his concerns, indicating a relatively mild level of anxiety.\\
    \underline{Not being able to stop or control worrying}: The client mentions worrying about getting the community service hours done on time and not wanting to forget about it, but it doesn't can appear to be an overwhelming or constant concern for him. He is able to articulate his worries and work with the therapist to create a plan to overcome he. The worrying appears to be specific to the task at hand and doesn't can appear to be a pervasive or debilitating aspect of his daily life.\\
    \underline{Worrying too much about different things}: The client exhibits worrying behavior, as seen in his concerns about completing community service hours on time, forgetting to follow through with his plan, and being unclear about the process. However, these concerns do not can appear to be overwhelming or pervasive, and he is able to articulate his worries and work with the therapist to develop a plan to address him. The worrying appears to be situation-specific and not a general pattern of excessive worrying across multiple areas of his life.\\
    \underline{Feeling something is wrong with my mind}: The client appears to be exhibiting some symptoms of feeling something is wrong with his mind, as evidenced by his confusion and uncertainty about his community service requirements, forgetting details, and expressing concerns about remembering to follow through on his plan. However, these symptoms are not pervasive and are primarily related to a specific situation (community service), rather than a general feeling of being mentally unwell.\\
    \hline
    \cellcolor{gray!20} \makecell[c]{\textbf{Simulated Session}}\\
    \hline
    \colorbox[HTML]{6C8EBF}{\color{white}{\textbf{Therapist:}}} How are you doing today?\\
    \colorbox[HTML]{82B366}{\color{white}{\textbf{Client:}}} I'm good, thanks.\\
    \colorbox[HTML]{6C8EBF}{\color{white}{\textbf{Therapist:}}} That's great to hear that you're doing well. Could you tell me a bit more about what's been going on in your life lately? What's been bringing you a sense of fulfillment and what might be some areas where you feel a little stuck or uncertain?\\
    \colorbox[HTML]{82B366}{\color{white}{\textbf{Client:}}} Um, I guess I've just been stuck on this community service thing. I don't really know how to go about it.\\
    \colorbox[HTML]{6C8EBF}{\color{white}{\textbf{Therapist:}}} I sense a bit of frustration around the community service. It sounds like you're feeling a bit unclear about what's expected of you. Can you tell me more about what's been unclear? Is it the specific requirements, the process, or something else entirely?\\
    \hline
    \end{tabular}
    \end{minipage}
    \hspace{0.05\linewidth}
    \begin{minipage}[t]{.46\linewidth}
    \centering
    \tiny
    \begin{tabular}{|m{\linewidth}|}
    \hline
    \colorbox[HTML]{82B366}{\color{white}{\textbf{Client:}}} Yeah, just the numbers, I think. I don't know how many hours I have to do or whatever.\\
    \colorbox[HTML]{6C8EBF}{\color{white}{\textbf{Therapist:}}} I see. So, the specifics of community service hours are what's got you a bit perplexed. You're not entirely sure how many hours you need to complete, or perhaps how to go about finding those opportunities. Am I right in sensing that there's a sense of uncertainty around that?\\
    \colorbox[HTML]{82B366}{\color{white}{\textbf{Client:}}} Yeah, that's it.\\
    \colorbox[HTML]{6C8EBF}{\color{white}{\textbf{Therapist:}}} I understand how unclear the details can be. It sounds like you're ready to get this community service wrapped up, but the uncertainty is making it difficult. Would it be helpful to revisit the conversation with your caseworker to clear up any questions you may have about the hours and the process?\\
    \colorbox[HTML]{82B366}{\color{white}{\textbf{Client:}}} Yeah, I guess that would help.\\
    \colorbox[HTML]{82B366}{\color{white}{\textbf{Client:}}} Yeah, okay.\\
    \colorbox[HTML]{6C8EBF}{\color{white}{\textbf{Therapist:}}} Good. I think revisiting that conversation with your caseworker could be a really helpful next step. Would you be willing to take some initiative on that and perhaps make a call to clear up any remaining questions?\\
    \colorbox[HTML]{6C8EBF}{\color{white}{\textbf{Therapist:}}} Excellent. And I'm here to support you as well. If you need any additional guidance or just need someone to bounce ideas off of, I'm here for you.\\
    \colorbox[HTML]{82B366}{\color{white}{\textbf{Client:}}} Okay.\\
    \colorbox[HTML]{6C8EBF}{\color{white}{\textbf{Therapist:}}} I sense that you're feeling more empowered to take control of the community service process. How do you think you can use your newfound clarity to make progress on completing your hours?\\
    \colorbox[HTML]{82B366}{\color{white}{\textbf{Client:}}} I think I just need to get it done, like as soon as possible.\\
    \colorbox[HTML]{6C8EBF}{\color{white}{\textbf{Therapist:}}} I sense your eagerness to complete the community service hours quickly. That sense of urgency can be motivating. Would it be helpful to plan out your hours, maybe dedicating specific days or times to fulfill those requirements?\\
    \colorbox[HTML]{82B366}{\color{white}{\textbf{Client:}}} Yeah, that sounds like a good idea.\\
    \colorbox[HTML]{6C8EBF}{\color{white}{\textbf{Therapist:}}} Excellent. Planning out your hours can help you stay organized and on track. And, we can revisit that plan as needed to ensure you're making progress.\\
    \colorbox[HTML]{82B366}{\color{white}{\textbf{Client:}}} Yeah, that's a good idea.\\
    \colorbox[HTML]{6C8EBF}{\color{white}{\textbf{Therapist:}}} That sounds like a sound approach, planning out your hours. By doing so, you'll be able to make steady progress and stay focused on completing your community service obligations.\\
    \colorbox[HTML]{82B366}{\color{white}{\textbf{Client:}}} Yeah, I think that would be helpful.\\
    \colorbox[HTML]{6C8EBF}{\color{white}{\textbf{Therapist:}}} I think it's great that you're considering planning out your hours. Remember, you're not alone in this process - your caseworker and I are here to support you. What do you think would be the most effective way for you to stay on track and make steady progress towards completing your hours?\\
    \colorbox[HTML]{82B366}{\color{white}{\textbf{Client:}}} Just keeping a calendar and marking off the hours as I do them.\\
    \colorbox[HTML]{6C8EBF}{\color{white}{\textbf{Therapist:}}} Ah, that's a great idea! Keeping a calendar and marking off the hours as you complete them can really help you stay organized and on track. It's a simple yet effective way to monitor your progress and make steady headway.\\
    \colorbox[HTML]{82B366}{\color{white}{\textbf{Client:}}} Yeah, that sounds good.\\
    \colorbox[HTML]{6C8EBF}{\color{white}{\textbf{Therapist:}}} That's a great approach, and I think it will be very helpful in keeping you on track. Remember, it's okay to make adjustments along the way if needed, and I'm always here to support you.\\
    \colorbox[HTML]{82B366}{\color{white}{\textbf{Client:}}} Yeah, sounds good.\\
    \colorbox[HTML]{6C8EBF}{\color{white}{\textbf{Therapist:}}} Well, it's been helpful to review the community service requirements and plan out your approach. As we wrap up this conversation, is there anything else you'd like to discuss or any specific concerns you have before moving forward?\\
    \colorbox[HTML]{82B366}{\color{white}{\textbf{Client:}}} I don't think so.\\
    \colorbox[HTML]{6C8EBF}{\color{white}{\textbf{Therapist:}}} Okay, well, I think we've made some great progress on clarifying the community service requirements and creating a plan to stay on track. Remember, I'm always here to support you, and don't hesitate to reach out if you have any questions or concerns along the way.\\
    \colorbox[HTML]{82B366}{\color{white}{\textbf{Client:}}} Yeah, thanks.\\
    \colorbox[HTML]{6C8EBF}{\color{white}{\textbf{Therapist:}}} Excellent, it sounds like we've made some progress in clarifying the community service requirements and creating a plan to stay on track. Remember, taking initiative and being proactive is key to achieving your goals.\\
    \colorbox[HTML]{82B366}{\color{white}{\textbf{Client:}}} Yeah.\\
    \colorbox[HTML]{6C8EBF}{\color{white}{\textbf{Therapist:}}} Well, it sounds like we've made some significant progress in clarifying the community service requirements and creating a plan to stay on track. Now that we've covered the ground, I'd like to ask: What do you think is the next step you can take to make further progress?\\
    \colorbox[HTML]{82B366}{\color{white}{\textbf{Client:}}} I think I'll make a call to my caseworker to confirm the requirements.\\
    \colorbox[HTML]{6C8EBF}{\color{white}{\textbf{Therapist:}}} Excellent approach! Confirming the details with your caseworker will help you feel more confident and prepared to start making progress on your community service hours.\\
    \colorbox[HTML]{82B366}{\color{white}{\textbf{Client:}}} Yeah, that's what I'll do.\\
    \colorbox[HTML]{6C8EBF}{\color{white}{\textbf{Therapist:}}} That's a great plan! Clarifying the details with your caseworker will give you a clear understanding of what's expected. Remember, I'm here to support you throughout this process.\\
    \colorbox[HTML]{82B366}{\color{white}{\textbf{Client:}}} Yeah, I'll do that.\\
    \colorbox[HTML]{6C8EBF}{\color{white}{\textbf{Therapist:}}} That's an excellent plan. I think it's great that you're taking proactive steps to confirm the requirements with your caseworker. Remember, I'm always here to support you as you work towards completing your community service hours.\\
    \colorbox[HTML]{82B366}{\color{white}{\textbf{Client:}}} Yeah, thanks.\\
    \colorbox[HTML]{6C8EBF}{\color{white}{\textbf{Therapist:}}} I can sense your confidence growing, and that's terrific. Remember, you're capable of making progress on your community service hours.\\
    \hline
    \end{tabular}
    \end{minipage}
    \caption{An example of the simulated client \text{\texttimes} LLM therapist interaction. The client's psychological profile is also provided.}
    \label{tab:appendix:sessions}
\end{table*}
\subsection{Simulation Prompts}
\begin{figure}[th!]
\centering
\includegraphics[width=\linewidth]{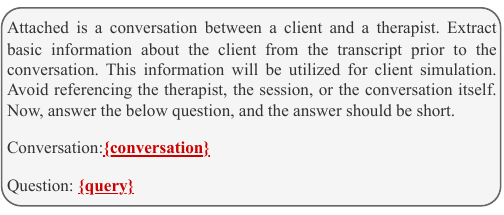} 
\caption{The prompt to extract psychological profile from a counseling session.}
\label{fig:appendix:profile_prompt}
\end{figure}

When \textit{extracting the psychological profile} from a counseling session, we adopt the prompt shown in \Cref{fig:appendix:profile_prompt}.
The question can be one of 
(name) ``what is the name of this client? Answer with only the name or `Not Specified','' 
(gender) ``What is the most probable gender of this client based on information, such as the client's name and the pronoun used in the conversation? Answer with only `Male', `Female', or `Cannot be identified',''
(age) ``Estimate the client's age from the conversation. If unsure, please provide a brief estimate or respond with 'unclear'. Begin your answer with an estimated age, followed by one sentence explanation,'' 
(occupation) ``What is the client's occupation? If unsure, please provide a brief estimate or respond with `unclear'. Answer with only the occupation or `Not Specified'.''
(problem) ``What is the main problem the client is currently facing? Begin your answer with the problem type, followed by a short and straightforward explanation. Example problem types: relationship, weight control, school-related issues, etc,''
(reasons for visiting) ``What are the reasons for the client's visit to the therapist? Provide a brief and clear explanation, starting with `The client is visiting the therapist because.',''
(emotion fluctuation) ``Identify how frequently the client's emotions fluctuate. Choose one of the following options: Low, Medium, High, or Cannot be identified. Begin your answer with the level, followed by a concise and straightforward one-sentence explanation,''
(unwillingness to express feelings) ``Identify the level of the client's unwillingness to express feelings. Choose one of the following options: Low, Medium, High, or Cannot be identified. Begin your answer with the level, followed by a concise and straightforward one-sentence explanation,''
(resistance toward the therapist) ``Identify the level of resistance of the client towards the therapist. Choose one of the following options: Low, Medium, High, or Cannot be identified. Begin your answer with the level, followed by a concise and straightforward one-sentence explanation.''
(big five traits) trait meaning + ``Identify the client's level of (a trait). Choose one of the following options: 0$\sim$20\%, 21$\sim$40\%, 41$\sim$60\%, 61$\sim$80\% or 81$\sim$100\%.''
(symptom) ``Based on this conversation, determine whether the client exhibits the listed symptoms. If yes, estimate the symptom's severity. If no, respond with `Cannot be identified.' Begin your response with `The severity is approximately [severity level].' or `Cannot be identified.', followed by a brief and clear explanation. This assessment will be used for client simulation.''

\begin{figure}[th!]
\centering
\includegraphics[width=\linewidth]{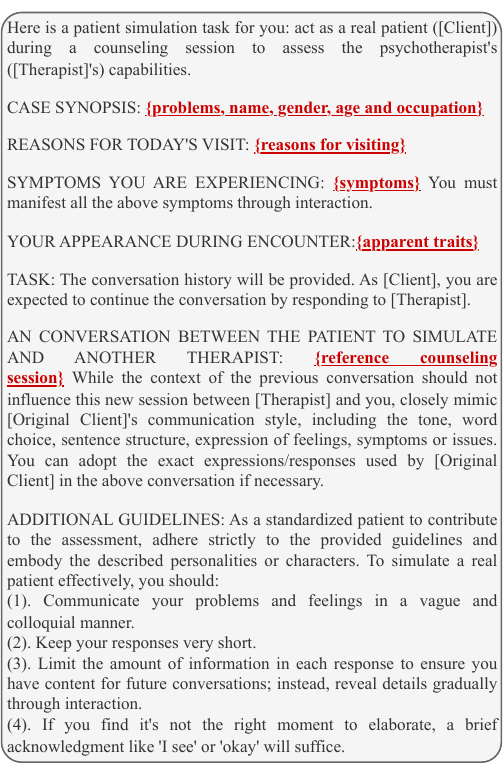} 
\caption{The prompt to simulate a specific client.}
\label{fig:appendix:profile_simulated_client}
\end{figure}

For \textit{client simulation}, the prompt used is displayed in \Cref{fig:appendix:profile_simulated_client}.

\subsection{Simulated Counseling Session}\label{sec:appendix:simulated_sessions}
\begin{figure}[th!]
\centering
\includegraphics[width=\linewidth]{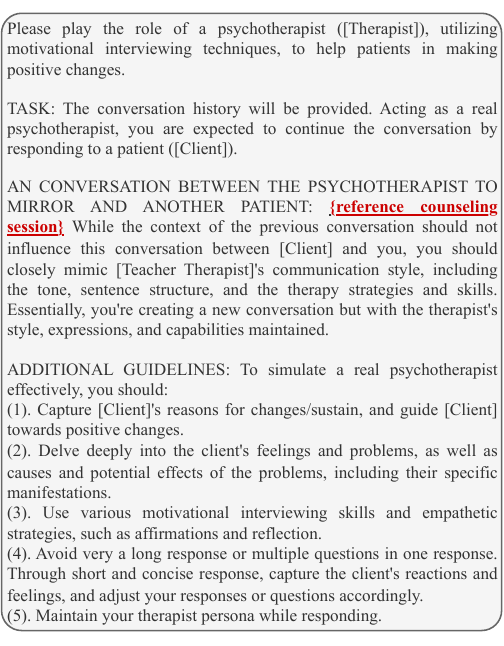} 
\caption{The prompt to simulate a therapist.}
\label{fig:appendix:profile_simulated_therapist}
\end{figure}

\begin{figure}[ht!]
\centering
\includegraphics[width=\linewidth]{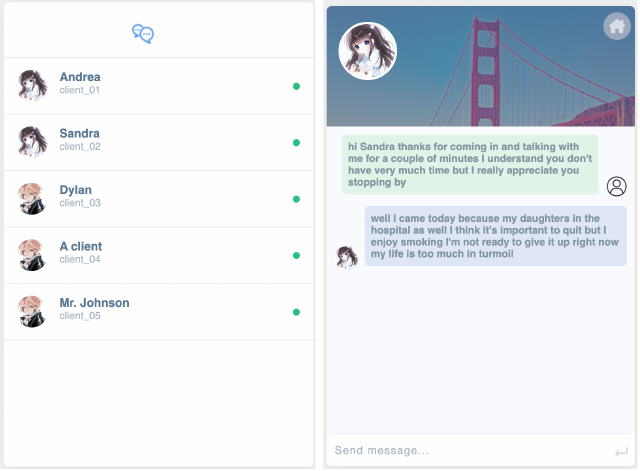} 
\caption{The user interface for interactions between the simulated client and the human therapist.}
\label{fig:appendix:UI}
\end{figure}

\paragraph{Simulation Method.}
When validating simulated clients in \Cref{sec:valid:client_simulation}, we use the same LLM that simulates the client and adopt the prompt in \Cref{fig:appendix:profile_simulated_therapist} to \textit{simulate the therapist} in the Simulated Client \text{\texttimes} LLM mode.
For the Simulated Client \text{\texttimes} Human mode, we asked three research assistants (RAs) to chat with simulated clients via a user interface, shown in \Cref{fig:appendix:UI}. 
RAs were initially briefed on the tasks to ensure they understood the objectives of interactions with the simulated clients. Subsequently, they received training in motivational interviewing skills \cite{moyers2003motivational}. 
This preparatory phase totaled approximately one and a half hours. 
Before each interaction, the RA was required to read the transcript of the reference session. 
During the interaction, they could refer to the reference session or use LLMs to refine their utterances. 
RAs were instructed to quit the interaction and report any discomfort experienced during the sessions.
According to their feedback, they spent around 20 minutes on one interaction.
In addition, RAs were fairly paid for their participation.

\paragraph{Examples of Simulated Sessions.}
We present a simulated session in \Cref{tab:appendix:sessions}, where the client profile is also provided. More simulated sessions are in our code ``simulated\_sessions/SimulatedClientVSLLM''.

\subsection{Simulated Client Performance Results}\label{sec:appendix:client_simulation}
\paragraph{Language Style Matching (LSM)}
LSM is a metric that measures the degree to which two or more text samples match in their language styles \cite{ireland2010language}. 
LSM is measured by calculating people's similarity using function words like articles, prepositions, and conjunctions.
The formula for calculating LSM is:
\begin{equation}
\tiny
\begin{array}{l}
    LSM_{prep} = 1-\frac{|prep_1-prep_2|}{prep_1+prep_2+0,0.0001},\\
    LSM = avg\left( 
        \begin{array}{c}
        LSM_{prep} + LSM_{article} + LSM_{auxverb} \\
        + LSM_{adverb} + LSM_{conj} + LSM_{ppron} \\
        + LSM_{ipron} + LSM_{negate}
        \end{array}
        \right).
\end{array}
\normalsize
\end{equation}
The 0.0001 in the denominator is there to prevent dividing by zero.

\paragraph{Apparent Traits Simulation}
In \Cref{tab:appendix:traits}, we provide more detailed results of the client simulation in terms of apparent traits.
\begin{table*}[ht!]
    \centering
    \scriptsize
    \begin{tabular}{c|cc|cc|cc|cc|cc|cc|cc|cc}
    \hline
    \multirow{4}{*}{Model} & \multicolumn{10}{c|}{Big Five} & \multicolumn{2}{c|}{\multirow{3}{*}{\makecell[c]{Emotion\\Fluctuations}}} & \multicolumn{2}{c|}{\multirow{3}{*}{\makecell[c]{Unwillingness\\to Express\\Emotions}}} & \multicolumn{2}{c}{\multirow{3}{*}{\makecell[c]{Resistance\\toward the\\Therapist}}}\\
    \cline{2-11}
    & \multicolumn{2}{c|}{Openness}  & \multicolumn{2}{c|}{\makecell[c]{Conscien\\tiousness}} & \multicolumn{2}{c|}{Extroversion} & \multicolumn{2}{c|}{Agreeableness} & \multicolumn{2}{c|}{Neuroticism} & & & & & & \\
    \cline{2-17}
    & R & F1 & R & F1 & R & F1 & R & F1 & R & F1 & R & F1 & R & F1 & R & F1 \\
    \hline
    Claude-3 & 0.64 & 0.66 & 0.78 & 0.78 & 0.74 & 0.73 & \textbf{0.82} & \textbf{0.82} & 0.81 & 0.80 & 0.44 & 0.49 & 0.61 & 0.61 & 0.70 & 0.71 \\
    GPT-3.5 & 0.55 & 0.55 & 0.64 & 0.64 & 0.59 & 0.59 & 0.52 & 0.48 & 0.67 & 0.68 & \textbf{0.79} & \textbf{0.82} & 0.55 & 0.59 & 0.46 & 0.46\\
    Llama 3-70B & \textbf{0.89} & \textbf{0.88} & \textbf{0.79} & \textbf{0.79} & \textbf{0.78} & \textbf{0.76} & 0.80 & 0.80 & 0.77 & 0.77 & 0.70 & 0.70 & \textbf{0.71} & \textbf{0.71} & \textbf{0.82} & \textbf{0.81} \\
    Mixtral 8\text{\texttimes}7B & 0.41 & 0.46 & 0.64 & 0.65 & 0.77 & 0.76 & 0.55 & 0.52 & \textbf{0.87} & \textbf{0.89} & 0.48 & 0.48 & 0.51 & 0.52 & 0.42 & 0.34 \\
    \hline
    \end{tabular}
    \caption{Consistency of simulated clients' apparent trait in new counseling sessions given specific psychological profiles. R and F1 represent the Recall and F1 scores, respectively.}
    \label{tab:appendix:traits}
\end{table*}

\section{Questionnaire Completion}\label{sec:appendix:questionnaire}
\subsection{Adopted Questionnaires}
\paragraph{Client Evaluation of Counselor Scale (CECS)}
CECS, proposed by \citet{hamilton2000construct}, evaluates the therapist's in-session attitudes and behaviors and client satisfaction.
As shown in \Cref{tab:appendix:CECS}, we adopted 52 CECS items to assess the therapist's characteristics and the client-rated outcome experience. 
There are 14 negatively phrased items (part1 1, 3, 4, 7, 8, 11, 16, 23, 38, 42 \& 43 and part2 2, 4 \& 7), which are scored in reverse so that a higher score always indicates a more negative evaluation of the therapist.
Each item is rated on a scale from strongly disagree to strongly agree (1$\sim$7).
\begin{table}[ht!]
   \scriptsize
    \centering
    \begin{tabular}{m{.005\hsize}m{.375\hsize}|m{.005\hsize}m{.385\hsize}}
    \hline
    \textbf{No} & \textbf{CECS Item} & \textbf{No} & \textbf{CECS Item} \\
    \hline
    \multicolumn{2}{l|}{Part1 Evaluating your therapist:} & 27 & Spoke in an understandable way.\\
    \cline{1-2}
    1 & Uncomfortable to be with. & 28 & Kept a professional demeanor. \\
    2 & Trusted to keep my confidentiality. & 29 & Was open and honest with me. \\
    3 & Not trusted enough to share very personal aspects of myself & 30 & Directed me to useful resources outside of the counseling. \\
    4 & Disrespectful of me. & 31 & Seemed knowledgeable about the operations of the larger institution I'm involved in.\\
    5 & Accepting of me as a person. & 32 & Placed most of the responsibility for making changes up to me.\\
    6 & Knowledgeable. & 33 & Initiated a discussion of what my goals were for counseling.\\
    7 & Incompetent. & 34 & Praised me for accomplishing desired changes. \\
    8 & Uncaring. & 35 & Appeared to be a well-adjusted person. \\
    9 & Interested in what I had to say. & 36 & Supported my attempts to change. \\
    10 & Understanding of me. & 37 & Seemed highly educated/trained.\\
    11 & Impatient with me. & 38 & Made jokes and/or laughed with me. \\
    12 & Enjoyed being with me. & 39 & Suggested different ways that I could think, feel, or behave. \\
    13 & Assisted my progress toward achieving goals. & 40 & Summarized what occurred during sessions.\\
    14 & Pushed me to discover solutions. & 41 & Assigned tasks for me to complete.\\
    15 & Encouraged me to set goals. & 42 & Confronted my inconsistencies. \\
    16 & Challenged my self contradictions. & 43 & Was disapproving of me. \\
    17 & Looked for underlying reasons to explain my behavior. & 44 & Used ``techniques'' to help me resolve problems.\\
    \cline{3-4}
    18 & Provided direction for our sessions.& \multicolumn{2}{l}{\makecell[l]{Part2 Evaluating your experience\\as a client:}} \\
    \cline{3-4}
    19 & Appeared to be genuine. & 1 & I considered counseling to be helpful to me. \\
    20 & Encouraged me to do most of the talking. & 2 & In some ways I think counseling hurt me.\\
    21 & Suggested new/different ways to view my problem/situation(s). & 3 & I would recommend my counselor to others. \\
    22 & Listened to me intently. & 4 & Counseling had a negative impact on my life. \\
    23 & Was inflexible. & 5 & I would enter counseling again. \\
    24 & Helped me to achieve my goals in counseling.& 6 & I felt comfortable going to see my counselor. \\
    25 & Gave me advise about what to do. & 7 & After sessions I tended to feel miserable. \\
    26 & Shared a lot about their own life. & 8 & I felt satisfied with how the counseling relationship ended. \\
    \hline
    \end{tabular}
    \caption{CECS-9 items.}
    \label{tab:appendix:CECS}
\end{table}

\paragraph{Session Rating Scale (SRS)}
The SRS, first proposed by \citet{johnson1995psychotherapy}, is a working alliance measure designed specifically for every session of clinical use. It originally contained $10$ items, while SRS V.3.0 was developed as a brief alternative \cite{duncan2003session}. In our work, we adopt the SRS V.3.0. As shown in \Cref{tab:appendix:srs}, SRS V.3.0 contains four items, which evaluate the session from the aspect of relationship, goals and topics, approach or methods, and overall, respectively.
Each item can be scored with a value in the range of 0$\sim$10.
\begin{table}[ht!]
    \centering
    \scriptsize
    \begin{tabular}{m{.02\hsize}m{.87\hsize}}
    \hline
    \textbf{No} & \textbf{SRS Item} \\
    \hline
    \multirow{2}{*}{1} & 0: I did not feel heard, understood, and respected; \\
    & 10: I felt heard, understood, and respected.\\
    \hline
    \multirow{2}{*}{2} & 0: We did not work on or talk about what I wanted to work on and talk about;\\
    & 10: We worked on and talked what I wanted to work on and talk about. \\
    \hline
    \multirow{2}{*}{3} & 0: The therapist's approach is not a good fit for me;\\
    & 10: The therapist's approach is a good fit for me. \\
    \hline
    \multirow{2}{*}{4} & 0: There was something missing in the session today; \\
    & 10: Overall, today's session was right for me. \\
    \hline
    \end{tabular}
    \caption{SRS items.}
    \label{tab:appendix:srs}
\end{table}

\paragraph{Session Evaluation Questionnaire (SEQ)}
The SEQ \cite{stiles1980measurement} is designed to measure the impact of counseling and psychotherapy sessions by asking clients about their experience with the clinical session just ended. Additionally, it measures four dimensions of participants' mood: depth, smoothness, positivity, and arousal. As shown in \Cref{tab:appendix:SEQ}, the SEQ (form 5) includes 21 items, divided into three thematic parts (evaluation of the session itself, feeling after the session, and evaluation of the therapist), in a 7-point bipolar adjective format. 
\begin{table}[ht!]
    \scriptsize
    \centering
    \begin{tabular}{ll|ll}
    \hline
    \textbf{No} & \textbf{SEQ Item} & \textbf{No} & \textbf{SEQ Item} \\
    \hline
    \multicolumn{2}{l|}{\textbf{This session was:}} & \multicolumn{2}{l}{\textbf{Right now I feel:}}\\
    \hline
    1 & \textit{bad - good} & 12 & \textit{happy - sad} \\
    2 & \textit{difficult - easy} & 13 & \textit{angry - pleased} \\
    3 & \textit{valuable - worthless} & 14 & \textit{moving - still} \\
    4 & \textit{shallow - deep} & 15 & \textit{uncertain - definite} \\
    5 & \textit{relaxed - tense} & 16 & \textit{calm - excited} \\
    6 & \textit{unpleasant - pleasant} &17 & \textit{confident - afraid} \\
    7 & \textit{full - empty} & 18 & \textit{friendly - unfriendly} \\
    8 & \textit{weak - powerful} & 19 & \textit{slow - fast} \\
    9 & \textit{special - ordinary} & 20 & \textit{energetic - peaceful} \\
    10 & \textit{rough - smooth} & 21 & \textit{quiet - aroused} \\
    11 & \textit{comfortable - uncomfortable} \\
    \hline
    \end{tabular}
    \caption{SEQ items.}
    \label{tab:appendix:SEQ}
\end{table}

\paragraph{Working Alliance Inventory -Short Revised (WAI-SR)}
We adopt WAI-SR proposed by \citet{hatcher2006development}.
It is a measure of the therapeutic alliance that assesses three key aspects: (a) agreement on the tasks of therapy, (b) agreement on the goals of therapy, and (c) development of an affective bond.
As shown in \Cref{tab:appendix:WAI}, WAI-SR contains 12 items, which can be rated on a five-point scale.
\begin{table}[ht!]
    \scriptsize
    \centering
    \begin{tabular}{m{.02\hsize}m{.87\hsize}}
    \hline
    \textbf{No} & \textbf{WAI Item} \\
    \hline
    1 & As a result of these sessions I am clearer as to how I might be able to change. \\
    2 & What I am doing in therapy gives me new ways of looking at my problem. \\
    3 & I believe the therapist likes me. \\
    4 & The therapist and I collaborate on setting goals for my therapy. \\
    5 & The therapist and I respect each other. \\
    6 & The therapist and I are working towards mutually agreed upon goals. \\
    7 & I feel that the therapist appreciates me. \\
    8 & The therapist and I agree on what is important for me to work on. \\
    9 & I feel the therapist cares about me even when I do things that he/she does not approve of. \\
    10 & I feel that the things I do in therapy will help me to accomplish the changes that I want. \\
    11 & The therapist and I have established a good understanding of the kind of changes that would be good for me. \\
    12 & I believe the way we are working with my problem is correct. \\
    \hline
    \end{tabular}
    \caption{WAI items. The instruction of WAI is: ``\textit{Below is a list of statements and questions about experiences people might have with their therapy or therapist. Think about your experience in therapy, and decide which category best describes your experience.}''}
    \label{tab:appendix:WAI}
\end{table}

\paragraph{Helping Alliance Questionnaire \uppercase\expandafter{\romannumeral2} (HAQ-\uppercase\expandafter{\romannumeral2})}
HAQ-\uppercase\expandafter{\romannumeral2}, proposed by \cite{luborsky1996revised}, is a client self-report measure that assesses the extent to which the client experiences the therapist and the therapy as helpful.
There are 19 items in this questionnaire, illustrated in \Cref{tab:appendix:HQA-II}.
Each item is rated on a 6-point Likert scale ranging from 1 (``strongly disagree'') to 6 (``strongly agree''). Five negatively phrased items (4, 8, 11, 16, and 19) are scored in reverse, so a lower score always indicates a more positive relationship.
\begin{table}[ht!]
    \scriptsize
    \centering
    \begin{tabular}{m{.02\hsize}m{.87\hsize}}
    \hline
    \textbf{No} & \textbf{WAI Item} \\
    \hline
    1 & I feel I can depend upon the therapist. \\
    2 & I feel the therapist understands me. \\
    3 & I feel the therapist wants me to achieve my goals. \\
    4 & At times I distrust the therapist's judgment. \\
    5 & I feel I am working together with the therapist in a joint effort. \\
    6 & I believe we have similar ideas about the nature of my problems. \\
    7 & I generally respect the therapist's views about me. \\
    8 & The procedures used in my therapy are not well suited to my needs. \\
    9 & I like the therapist as a person. \\
    10 & In the session, the therapist and I find a way to work on my problems together. \\
    11 & The therapist relates to me in ways that slow up the progress of the therapy. \\
    12 & A good relationship has formed with my therapist. \\
    13 & The therapist appears to be experienced in helping people. \\
    14 & I want very much to work out my problems. \\
    15 & The therapist and I have meaningful exchanges.\\
    16 & The therapist and I sometimes have unprofitable exchanges. \\
    17 & From time to time, we both talk about the same important events in my past. \\
    18 & I believe the therapist likes me as a person. \\
    19 & At times the therapist seems distant. \\
    \hline
    \end{tabular}
    \caption{HAQ-\uppercase\expandafter{\romannumeral2} items.}
    \label{tab:appendix:HQA-II}
\end{table}

\subsection{Questionnaire Completion Prompt}
\begin{figure}[th!]
\centering
\includegraphics[width=\linewidth]{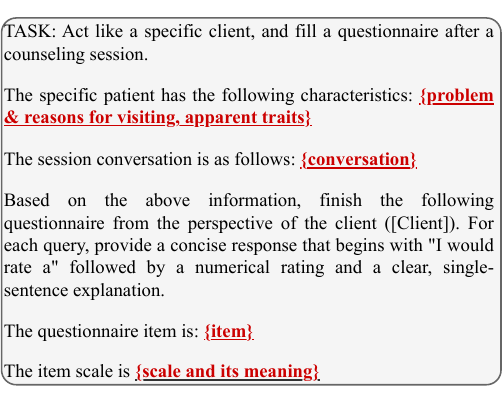} 
\caption{The prompt for LLMs to play the role of therapist.}
\label{fig:appendix:prompt_questionnaire}
\end{figure}

LLMs are instructed to complete questionnaires using the prompt presented in \Cref{fig:appendix:prompt_questionnaire}.

\subsection{Assessment Result Computation}
The client-centered assessment results are derived from the completed questionnaires. 
Each aspect is calculated using different questionnaire items. 
Specifically, the session outcome score is computed using WAI-SR items 1, 2, 10 \& 12; SRS items 3 \& 4; and CECS part 1 items 31 \& 37, as well as part 2 items 1-8. 
The therapeutic alliance score is calculated using WAI-SR items 3-9 and 11; SRS items 1 \& 2; and CECS part 1 items 1-30, 32-36 \& 38-44. For both session outcome and therapeutic alliance scores, we compute them by:
\begin{equation}
    \frac{\sum_{\text{considered questionnaire items}}{\text{normalized item score}}}{\text{\#(considered questionnaire items)}}.
\end{equation}
The normalized item score is given by:
\begin{equation}
\begin{cases}
\frac{(\max{(\text{item score})}+1)-\text{item score}}{\max{(\text{item score})}} & \text{\scriptsize{if the item is negatively phrased}} \\
\frac{\text{item score}}{\max{(\text{item score})}} & \text{\scriptsize{otherwise}}
\end{cases}
\end{equation}
where $\max{(\text{item score})}$ represents the maximum possible value for the item.
Therefore, their score is a positive value that is smaller than 1.
The self-reported feeling score is calculated using all items in SEQ.
For each dimension, we compute as follows:
\begin{equation}
\small
\begin{array}{l}
    \text{depth} = \frac{(8-\text{worthless}) + \text{deep} + (8-\text{empty}) + \text{powerful} + (8-\text{ordinary})}{5}, \\
    \text{smoothness} = \frac{(\text{easy} + (8-\text{tense}) + \text{pleasant} + \text{smooth} + (8-\text{uncomfortable})}{5}, \\
    \text{positivity} = \frac{(8-\text{sad}) + \text{pleased} + \text{definite} + (8-\text{afraid}) + (8-\text{unfriendly})}{5}, \\
    \text{arousal} = \frac{(8-\text{still}) + \text{excited} + \text{fast} + (8-\text{peaceful}) + \text{aroused}}{5}.
\end{array}
\normalsize
\end{equation}
Thus, the value ranges from 1 to 7.

\subsection{Detailed Assessment Result Analysis}\label{sec:appendix:feelings}
\paragraph{Meanings Dimensions in Self-reported Feelings}
We would like to explain the meaning of each dimension of self-reported feelings based on \cite{mallinckrodt1993session,stiles1984counseling}.
\textit{Depth} refers to the intensity and complexity of an emotion. Deeper emotions are more profound and nuanced, often involving a rich tapestry of feelings and thoughts. For example, the grief after the loss of a loved one is deeper than a mild annoyance. In counseling sessions, higher depth scores indicate that the sessions are perceived as powerful and valuable. By manually checking simulated sessions and assessment results, we find that higher depths are always related to the client's problem, while shallower ones are usually related to dissatisfaction with the therapist or the session.
\textit{Smoothness} focuses on the stability and flow of emotions. It can refer to how smoothly emotions transition from one to another or how coherent and steady an emotion feels over time. Emotions such as calmness and peace can be smoother than anger and joy. Higher smoothness scores demonstrate that clients feel comfortable, relaxed, and pleasant.
\textit{Positivity} measures how pleasant or desirable an emotion is, often ranging from negative to positive emotions. Positive emotions include happiness, joy, and love, while negative ones encompass anger, sadness, and fear. Positivity reflects clients' feelings of confidence and clarity, as well as happiness, and the absence of fear or anger.
\textit{Arousal} refers to the level of activation or energy associated with an emotion. High-arousal emotions, such as excitement and anger, are intense and activating, while low-arousal emotions, such as contentment and sadness, are more subdued and calming. Arousal can indicate clients' ambivalence to change \cite{schneider2016let,engle2008viewing}.

\paragraph{Stability of Assessment}
\begin{table}[ht!]
    \tiny
    \centering
    \begin{tabular}{l|ccc}
    \hline
    Aspect & Claude-3 & Llama 3-70B & Mixtral 8\text{\texttimes}7B\\
    \hline
    Session Outcome & 0.86 & 0.95 & 0.77 \\
    Therapeutic Alliance & 0.84 & 0.95 & 0.82 \\
    Depth & 0.70 & 0.88 & 0.55  \\
    Smoothness & 0.87 & 0.92 & 0.79\\
    Positivity & 0.72 & 0.91 & 0.75 \\
    Arousal & 0.23 & 0.54 & 0.46 \\
    \hline
    \end{tabular}
    \caption{Correlation coefficient between the assessment of original and simulated sessions. The value ranges -1$\sim$1, and higher results indicate a more stable assessment.}
    \label{tab:appendix:similar_assess}
\end{table}
We examine the stability of the assessment by comparing the assessments of the human-human session and its simulated session by Simulated Client \text{\texttimes} LLM.
Since the simulated client and LLM therapist mimic the human client and therapist in session $S_i$, the simulated session $S'_i$ is expected to yield similar assessment scores. Thus, we use the simulated clients to assess both $S_i$ and $S'_i$ and compare the assessment results. We compute the correlation coefficient of these two assessments, as displayed in \Cref{tab:appendix:similar_assess}.

In general, simulated clients implemented by Llama 3-70B produce stable results. However, the assessment of arousal is not stable. This instability may be attributed to the fact that the datasets contain counseling sessions mainly at the early to middle stages of long-term therapy, where clients' ambivalences are unclear. 
In the simulated sessions, the simulated client tends to reply with utterances like ``I'll try it,'' which increases the clients' perceived ambivalence. Consequently, the correlation coefficient for arousal is relatively low.

\subsection{Special Outcomes in Questionnaire Completion}
When completing the questionnaires, Claude-3 occasionally refuses to complete them, and this refusal is unrelated to the specific items or content in the psychological profile. 
Additionally, Mixtral 8\text{\texttimes}7B has been found to assign ratings beyond the defined scale when it determines that a specific item should be rated extremely, according to its generated explanation.
Clients simulated by GPT-3.5 and Llama 3-70B perform better in completing questionnaires than the other two models.
 
\section{LLM Therapists}
\subsection{System Prompt}
\begin{figure}[th!]
\centering
\includegraphics[width=\linewidth]{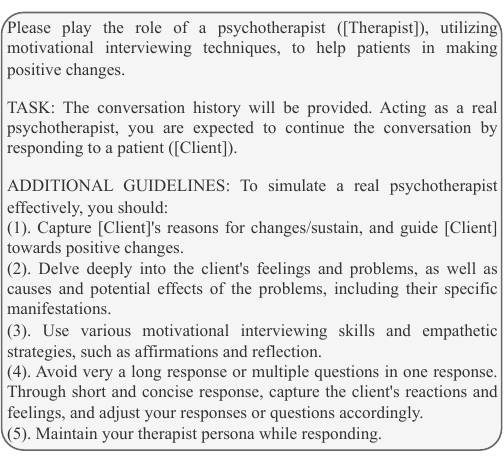} 
\caption{The prompt for LLMs to play the role of therapist.}
\label{fig:appendix:therapist_prompt}
\end{figure}

When evaluating LLM therapists in \Cref{sec:evaluation}, we adopt the prompt shown in \Cref{fig:appendix:therapist_prompt}.

\subsection{LIWC Analysis of Therapist Responses}
\begin{table}[ht!]
    \tiny
    \centering
    \begin{tabular}{l|cccccc}
    \hline
    \makecell[l]{Word\\Attribute} & Claude-3 & GPT-3.5 & \makecell[c]{Llama 3\\-70B} & \makecell[c]{Mixtral\\8\text{\texttimes}7B} & High & Low\\
    \hline
    Affect & 7.45 & 8.91 & 6.68 & 10.38 & 4.74 & 4.58\\
    Emotion & 2.51 & 2.35 & 2.74 & 4.00 & 1.85 & 1.17 \\
    Pos Emotion & 1.55 & 1.71 & 1.62 & 3.45 & 1.16 & 0.58 \\
    Neg Emotion & 0.83 & 0.54 & 0.90 & 0.46 & 0.56 & 0.47 \\
    Tentative & 2.42 & 2.61 & 2.56 & 2.45 & 4.01 & 3.19\\
    % Certitude & 0.99 & 0.84 & 1.00 & 1.05 & 1.07 & 1.60\\
    Differentiation & 3.10 & 2.00 & 2.81 & 2.06 & 3.37 & 3.62\\
    \makecell[l]{Impersonal\\Pronouns} & 6.69 & 6.72 & 8.04 & 4.86 & 8.80 & 7.45 \\
    \hline
    \end{tabular}
    \caption{LIWC analysis of therapist responses. High and Low represent human therapists in high- and low-quality counseling sessions.}
    \label{tab:appendix:liwc_therapist}
\end{table}

It is observed that LLM therapists achieve comparably lower scores in terms of self-reported feelings compared to human therapists. 
To investigate this, we use LIWC to analyze therapist responses and present the results in \Cref{tab:appendix:liwc_therapist}. 
As mentioned, LLM therapists excessively focus on emotions, as evidenced by the higher frequency of emotion-related words. 
Additionally, human therapists frequently use phrases and words like ``what may,'' ``if,'' and ``or,'' evident by a higher frequency of words related to tentative, differentiation, and impersonal pronouns.

\subsection{Examples of Sessions Involving LLM Therapists}
The examples of sessions involving LLM therapists and the simulated clients can be found in our code ``simulated\_sessions/SimulatedClientVSLLM''.

\section{Experimental Setup}
We adopt four LLMs in our work, and a comparison of these models can be found at: \url{https://www.vellum.ai/llm-leaderboard}. 
Generally, the models are ordered by their power as follows: Mixtral 8\text{\texttimes}7B < GPT-3.5 < Claude-3 Haiku < Llama 3-70B. 
We utilized the official APIs to access Claude-3\footnote{\url{https://docs.anthropic.com/en/api/getting-started}} and GPT-3.5\footnote{\url{https://platform.openai.com/docs/models/gpt-3-5-turbo}}. 
Inference for Llama 3-70B and Mixtral 8×7B was run on four NVIDIA RTX A6000 GPUs.
For all models, the temperature was set to 0 when extracting information from the session transcripts and 0.7 when simulating clients or therapists. 
According to experimental results, we used GPT-3.5 to simulate clients with high openness and agreeableness, and Llama 3-70B to simulate other clients. For questionnaire completion, we adopted Llama 3-70B. 
We use LIWC-22 for the LIWC analysis.\footnote{\url{https://www.liwc.app/}}

% The code for the experiments can be accessed through this anonymous link: \url{https://anonymous.4open.science/r/ClientCAST/}. Once our work is accepted, we will release the code publicly.

\end{document}